\documentclass[journal]{IEEEtran}
\IEEEoverridecommandlockouts
\usepackage{cite}
\usepackage{amsmath,amssymb,amsfonts}

\usepackage{moreverb,url}

\usepackage[colorlinks,bookmarksopen,bookmarksnumbered,citecolor=red,urlcolor=red]{hyperref}
\usepackage{algorithmic}
\usepackage{graphicx}
\usepackage{textcomp}
\usepackage{tabularray}
\usepackage[normalem]{ulem}
\usepackage{makecell}
\usepackage{array}
\usepackage{colortbl}
\usepackage{multirow}
\usepackage{hhline}

\usepackage[table,xcdraw]{xcolor}
\usepackage{mathtools}

\def\BibTeX{{\rm B\kern-.05em{\sc i\kern-.025em b}\kern-.08em
    T\kern-.1667em\lower.7ex\hbox{E}\kern-.125emX}}

\newcommand{\bzero}{\ensuremath{\mathbf{0}}}
\newcommand{\real}{\mathbb{R}}
\newcommand{\bH}{\ensuremath{\mathbf{H}}}
\newcommand{\bM}{\ensuremath{\mathbf{M}}}
\newcommand{\bA}{\ensuremath{\mathbf{A}}}
\newcommand{\bK}{\ensuremath{\mathbf{K}}}
\newcommand{\bE}{\ensuremath{\mathbf{E}}}
\newcommand{\bC}{\ensuremath{\mathbf{C}}}

\newcommand{\bu}{\ensuremath{\mathbf{u}}}
\newcommand{\bdu}{\ensuremath{\mathbf{\delta u}}}
\newcommand{\bU}{\ensuremath{\mathbf{U}}}
\newcommand{\bcU}{\pmb{\mathcal U}}
\newcommand{\bcV}{\pmb{\mathcal V}}

\newcommand{\bB}{\ensuremath{\mathbf{B}}}
\newcommand{\bD}{\ensuremath{\mathbf{D}}}
\newcommand{\bW}{\ensuremath{\mathbf{W}}}
\newcommand{\bY}{\ensuremath{\mathbf{Y}}}
\newcommand{\bL}{\ensuremath{\mathbf{L}}}

\newcommand{\tbA}{\tilde{\bA}}
\newcommand{\tbB}{\tilde{\bB}}

\newcommand{\tbC}{\tilde{\bC}}
\newcommand{\tbD}{\tilde{\bD}}

\newcommand{\hbK}{\widehat{\bK}}
\newcommand{\hbC}{\widehat{\bC}}
\newcommand{\hbB}{\widehat{\bB}}
\newcommand{\Ir}{\mathbf{I}_r}
\newcommand{\Idm}{\mathbf{I}_m}

\newcommand{\by}{\ensuremath{\mathbf{y}}}

\newcommand{\bx}{\ensuremath{\mathbf{x}}}

\newcommand{\bX}{\ensuremath{\mathbf{X}}}
\newcommand{\hbx}{\widehat{\bx}}
\newcommand{\hbX}{\widehat{\bX}}

\begin{document}

\title{Dynamic Shape Control of Soft Robots Enabled by Data-Driven Model Reduction}
\author{
\IEEEauthorblockN{Iman Adibnazari$^1$, Harsh Sharma$^2$, Myungsun Park$^1$, Jacobo Cervera-Torralba$^1$, Boris~Kramer$^1$,~Michael~T.~Tolley$^{1, 3}$}

\IEEEauthorblockA{\textit{$^1$Department of Mechanical and Aerospace Engineering, University of California San Diego, CA, USA}
}
\IEEEauthorblockA{\textit{$^2$Department of Mechanical Engineering, University of Wisconsin-Madison, WI, USA}
}
\IEEEauthorblockA{\textit{$^3$Materials Science and Engineering Program, University of California San Diego, CA, USA}
}
}
\maketitle

\begin{abstract}
Soft robots have shown immense promise in settings where they can leverage dynamic control of their entire bodies. However, effective dynamic shape control requires a controller that accounts for the robot's high-dimensional dynamics---a challenge exacerbated by a lack of general-purpose tools for modeling soft robots amenably for control. In this work, we conduct a comparative study of data-driven model reduction techniques for generating linear models amendable to dynamic shape control. We focus on three methods---the eigensystem realization algorithm, dynamic mode decomposition with control, and the Lagrangian operator inference (LOpInf) method. Using each class of model, we explored their efficacy in model predictive control policies for the dynamic shape control of a simulated eel-inspired soft robot in three experiments: 1) tracking simulated reference trajectories guaranteed to be feasible, 2) tracking reference trajectories generated from a biological model of eel kinematics, and 3) tracking reference trajectories generated by a reduced-scale physical analog. In all experiments, the LOpInf-based policies generated lower tracking errors than policies based on other models.
\end{abstract}

\section{Introduction}
Soft robotic systems have proven valuable in many settings, from enabling marine exploration \cite{aracri_soft_2021} and novel medical technologies \cite{cianchetti_biomedical_2018}, to providing crucial insights into the biomechanics and ethology of biological organisms \cite{ijspeert_biorobotics_2014}.
The benefits of soft robots often stem from the affordances provided by the compliance of the soft materials comprising them. For example, material compliance can be used to provide guarantees of mechanical safety in the event of collisions between a soft robot and delicate objects or environments (e.g., human tissue, coral reefs) \cite{cianchetti_biomedical_2018, youssef_underwater_2022}. 

While the affordances provided by compliant materials and structures are central to the use of soft robotic systems, this compliance simultaneously poses key challenges to their development and broader adoption. One key challenge is a current lack of effective and general-purpose algorithmic tools for controlling soft systems  \cite{george_thuruthel_control_2018}. General-purpose tools for controlling soft robots are challenging to develop because soft systems generally have highly underactuated continuum dynamics (i.e., the robot's structure has infinite degrees of freedom and finite degrees of actuation) and suffer from myriad nonlinear effects, including material nonlinearities (e.g., hysteresis, viscoelastic effects) and geometric nonlinearities under loading. 
\begin{figure}[tb]
    \centerline{\includegraphics[width=3.25in]{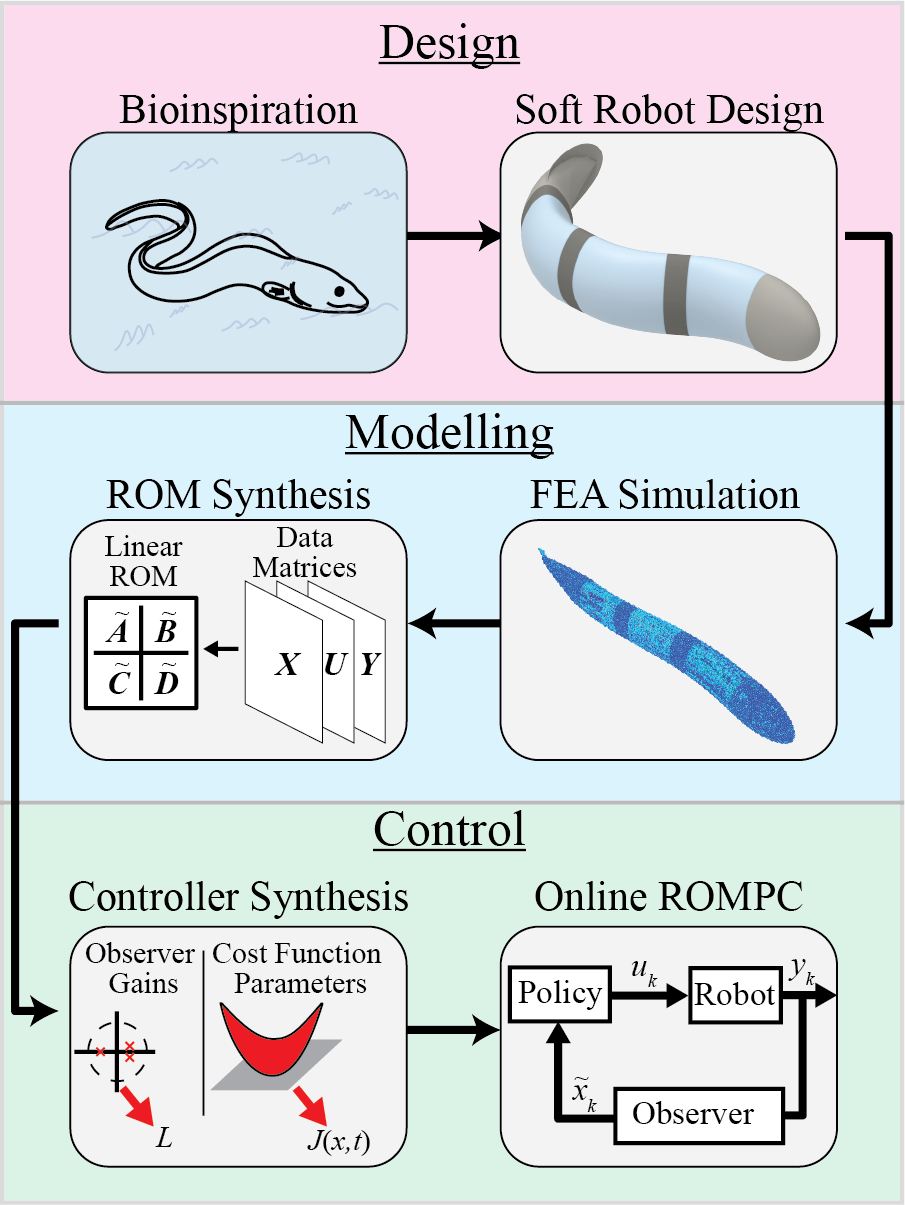}}
    \caption{Process flow for synthesizing a reduced-order model predictive controller for dynamic shape control of a soft robot. Our proposed process spans three stages---design (top), modeling (middle), and control (bottom)---that we defined to be broadly applicable to a variety of soft robots. In this work, we focus on an anguilliform-inspired soft robot, designed through a process of bioinspiration (top). Using data generated from a high-fidelity finite element simulation of the robot, we explored various techniques for data-driven model reduction to generate linear reduced-order models that are amenable to control. We then used these ROMs to construct a closed-loop state observer and reduced-order model predictive controller.}
    \label{fig1}
\end{figure}

\subsection{Literature Review}
Prior work on the dynamic control of soft robots has most frequently focused on the control of a single point of interest---typically the position of an end-effector~\cite{haggerty_control_2023, Bruder2019a, george_thuruthel_control_2018}.
However, there is increasing interest in dynamically controlling a soft robot's full continuum structure in both manipulation and locomotion tasks~\cite{george_thuruthel_control_2018, haggerty_control_2023, singh_controlling_2023};  snakes~\cite{onal_autonomous_2013,arachchige_soft_2021,liu_reinforcement_2023}, fish~\cite{nguyen_evaluation_2021,nguyen_anguilliform_2022}, cephalopods~\cite{george_thuruthel_control_2018, xie_octopus-inspired_2023}, and mammals with prehensile tails~\cite{doerfler_hybrid_2023} frequently serve as inspiration for the benefits of dynamically controlled hybrid structures that combine rigid and soft materials. For these robots, dynamic shape control can be used to enable improved (e.g., more efficient) operation as well as richer interactions with their respective environments. These benefits sit in contrast to those provided by systems that use quasi-static shape control for locomotion tasks (e.g., quasi-static inchworm~\cite{zhang_inchworm_2022} and walking gaits~\cite{drotman_electronics-free_2021} used to improve locomotion stability), and manipulation tasks in contexts where dynamic movement is undesirable (e.g., minimally invasive surgery~\cite{xu_shape_2024}). 

The dynamic shape control of a soft robot requires a control system that effectively accounts for the continuum mechanics governing the robot's behavior under finite degrees of actuation. However, past work on model-based control of soft robots has shown that the selection of a model frequently poses a tradeoff between model fidelity and model parsimony---generally affecting controller precision and speed \cite{haggerty_control_2023, della_santina_model-based_2023}. To address this challenge, four dominant paradigms have emerged for dynamic control of soft robotic systems: a) Model-based control with reduced physics models b) Reinforcement learning, c) Model-based control with numerical finite element simulations and intrusive reduced-order models thereof, and d) Linear control with models generated via extended dynamic mode decomposition methods. We will briefly survey these methods next.

\paragraph{Model-based Control with Reduced Physics Models}
The first of these paradigms emphasizes the use of reduced physics models in developing control systems for a given soft robot. Common  examples of reduced physics models can be found in  piecewise constant strain models of soft manipulators and models based on Cosserat rod theory \cite{armanini_soft_2023,li_piecewise_2023,till_real-time_2019,doroudchi_configuration_2021}. Reduced physics models frequently assume quasi-static, discrete-state, and/or fully actuated behavior from a soft robot~\cite{george_thuruthel_control_2018,armanini_soft_2023}, thereby diminishing, or even entirely removing, various dynamic phenomena that are otherwise realized by the physical system. 
In a control setting, the benefits of using reduced physics models lies in their interpretability, enabling the development of control systems with provable guarantees (e.g., stability, optimality, robustness) with respect to the underlying model~\cite{george_thuruthel_control_2018}. For example, research on piecewise constant strain models has matured to the extent where these models admit many of the analytical and computational tools that have become common place for traditional, rigid-body systems \cite{armanini_soft_2023, katzschmann_dynamic_2019, till_real-time_2019}. 
However, the benefits of controllers based on reduced physics models are predicated on the accuracy of the assumptions underlying the model, which can degrade significantly as the complexity of the soft robot increases. Furthermore, constructing new reduced physics models requires significant expertise, is labor intensive, and results in a model that typically only applies to a single class of soft robot.

\paragraph{Reinforcement Learning}
A second prominent paradigm for soft robot control circumvents the need for explicitly modeling the robot's dynamics by leveraging various reinforcement learning techniques\cite{george_thuruthel_control_2018,thuruthel_model-based_2019,wang_learn_2022,liu_reinforcement_2023}. This approach has been employed to control soft manipulators \cite{centurelli_closed-loop_2022,thuruthel_model-based_2019}, soft swimming robots \cite{wang_learn_2022, li_deep_2021}, and soft walking robots \cite{ji_synthesizing_2022}, often producing competent control policies that implicitly learn a model of the dynamics of the robot by training on data taken from an experimental or simulated system. However, the infinite dimensionality of soft robots implies requirements for large amounts of training data, especially as the number of inputs or outputs in the system increases~\cite{bhagat_deep_2019,kim_review_2021}. Moreover, control policies produced through this approach often provide few, if any, guarantees on stability, performance, and safety---factors crucial to operation in safety-critical environments like marine environments or around human bodies~\cite{bhagat_deep_2019}. 

\paragraph{Model-based Control with Numerical Finite Element Simulations and Intrusive Reduced Order Models}
Another paradigm for the control of soft robots focuses on developing model-based control policies around fast finite element simulations of a given system~\cite{della_santina_model-based_2023,schegg2022review}. A primary benefit of this approach lies it in being broadly applicable to nearly arbitrary soft robots while also often providing accurate control of the system as a product of using high-fidelity models based on finite element methods (FEM)~\cite{goury2018fast}. However, a fundamental challenge in using FEM-based models lies in their computational costs~\cite{schegg2022review}. One approach to addressing this challenge has focused on developing real-time FEM simulation tools for the control the soft robots with sufficiently low-dimensional FEM models~\cite{duriez_control_2013,coevoet2017optimization}. For systems with higher-dimensional models however, various methods for projection-based model order reduction have been proposed to generate fast and low-dimensional numerical models suitable for use with linear control~\cite{katzschmann2019dynamically, thieffry2018control, tonkens2021soft, coevoet2017optimization, goury2021real}. The techniques developed in these examples were shown to provide significant speed-ups to FEM simulations of soft robotic systems, enabling these models to be used in online control. Additionally, model order reduction and hyperreduction methods have also been shown to handle challenging settings where the robot might exhibit material and geometric nonlinearities as well as contacts with the environment \cite{coevoet2017optimization, goury2021real}. However, these methods are ultimately \textit{intrusive} in the sense that they require the finite element simulator to provide access to the high-dimensional operators of the full-order model (FOM). This requirement poses challenges in settings where these operators are inaccessible and in systems that have limited computational resources, as is commonly the case for mobile robots. 

\paragraph{Linear Control with Models Generated via Extended Dynamic Mode Decomposition Methods:}
The most recent paradigm to emerge for dynamic control of soft robots combines well-established tools for linear control with data-driven linear models generated through various extended dynamic mode decomposition (eDMD) methods \cite{ haggerty_modeling_2020, bruder_data-driven_2021,Bruder2019a, castano2020control}. These eDMD-based models have been extensively explored and validated through the lens of Koopman operator theory~\cite{mezic_koopman_2021, brunton_modern_2022}, wherein nonlinear lifting transformations are used to construct linear representations of otherwise nonlinear system dynamics. In practice, these transformations are applied to sensor data from a given robot, and the accuracy of learned models (and resulting quality of control) is often highly sensitive to one's selected library of lifting transformations~\cite{otto_linearly_2019, williams2015data}. A common strategy for producing eDMD-based linear models involves generating a large library of lifting transformation from a given family of basis functions (e.g., polynomials, time-delay functions). However, simply increasing the number observables does not necessarily improve accuracy and the dimension of the learned model can scale quickly with the the number of sensors used during operation and number of observable functions used for model synthesis~\cite{otto_linearly_2019}. 

When applied to the control of soft robots, a primary appeal of eDMD-based modeling and control lies in the ease of model synthesis directly from system data and the enforced linear structure of the resulting model. This linear structure enables the use of mature control tools such as linear quadratic regulator (LQR) control~\cite{haggerty_control_2023}, and linear model predictive control (MPC)~\cite{bruder_data-driven_2021,Bruder2019a}---wherein constraints on the input, output, and state of the system can be enforced during operation.  

\vspace{5pt}
MPC frameworks are particularly well-suited for dynamic shape control of soft robots as they allow for explicit consideration of the robot's dynamics, operating constraints, and external forcing from disturbances and interactions~\cite{borrelli_predictive_2017}. However, the manual selection of lifting transformations central to the control of soft robots with eDMD-based models often ignores the various physics and system-theoretic structure inherent to soft robot operation---structure that can be leveraged in a dynamic shape control setting. Alternative classes of \textit{non-intrusive} data-driven model reduction techniques can enforce such underlying structure in fast and low-dimensional models synthesized from collected data. For example, some system identification techniques represent dynamics with guarantees on the controllability or observability of the resulting system~\cite{juang_eigensystem_1985}, while others estimate models that preserve second-order mechanical structure of the system~\cite{sharma2024preserving, sharma2024data,sharma2024lagrangian}. 

\subsection{Contributions}
In this work, we explore the space of data-driven model reduction techniques to be used on model-predictive dynamic shape control for soft robots. We focus on the dynamic shape control of a simulated soft anguilliform robot (Fig.~\ref{fig1}), motivated by the facts that anguilliform swimmers are among the most efficient natural swimmers, that their motion is highly dynamic in nature, and that their motion requires the careful coordination of the entire body of the fish to effectively produce thrust~\cite{tytell_hydrodynamicsI_2004,tytell_hydrodynamicsII_2004, hess2024continuum}. Therefore, a robot capable of similar motion requires effective dynamic shape control of its body. 

Our contributions in this work are: 1) a generalizable pipeline for the modeling and dynamic shape control of soft robots enabled by data-driven linear model reduction, illustrated in Figure~\ref{fig1}; 2) a comparative study of three non-intrusive model reduction methods, each assessed in estimation and control settings; 3) a high-fidelity, finite element structural simulation of a fluidically driven anguilliform soft robot to serve as a testbench for dynamic shape controllers; and 4) a freely available large-scale dataset of the dynamic behavior of the aforementioned simulated soft robot to facilitate further work on data-driven modeling of soft robotic systems~\cite{Adibnazari2025Dataset}. 
Our simulated testbench only considers the structural mechanics of the robot under the forcing of pressure inputs that we assume propagate at much faster timescales than structural deformations. We made this choice because of the limited available software tooling and immense computational expense related to simulating three-dimensional fluid-structure interaction (FSI) for soft robots. However, by focusing on structural mechanics, our comparative study addresses the challenges of dynamic shape control that are most relevant to the majority of soft robotic systems and accomplishes a necessary first step in effectively controlling soft robots that are subject to significant effects from fluid-structure interaction.

\subsection{Paper Outline}
We begin in Section~\ref{sec:SoftRobotSim} by introducing the simulated anguilliform soft robot used as a testbench throughout this work followed by our formulation for reduced-order model predictive control (ROMPC) in Section~\ref{sec:rompcFormulation}. We introduce the model reduction techniques used in our comparative study in Section~\ref{sec:DataDrivenROMs} and discuss their training and performance within an estimation setting in Section~\ref{sec:romTraining}. We then compare these methods in the context of dynamic shape control in Section~\ref{sec:rompcTests} on reference trajectories that are dynamically feasible. Following this comparison, we assess the performance of ROM-based controllers based on each model reduction method in two sets of control experiments wherein reference trajectories are generated from data collected from a physical system. In Section~\ref{sec:controlExperiment1}, we use our control scheme to track reference trajectories modeled after the gaits of anguilliform fish and in Section~\ref{sec:controlExperiment2} we track reference trajectories generated by a reduced-scale physical analog of our simulated platform. Finally, we discuss our results in Section~\ref{sec:Discussion} and present conclusions and potential avenues for future work in Section~\ref{sec:Conclusion}. 

\section{Anguilliform Soft Robot Design and Simulation}
\label{sec:SoftRobotSim}

To center this work in a setting where dynamic shape control is necessary for successful operation, we constructed a large-scale, simulated testbench of the eel-inspired soft robot developed in \cite{cervera-torralba_lost-core_2024} and \cite{park_analysis_2025} that models the dynamics of the robot using an open-source finite element simulation tool (SOFA, SOFA Consortium)~\cite{payan_sofa_2012, coevoet_software_2017}. In this section, we briefly review the design and operation of the robot being simulated in Section~\ref{subsec:SoftRobotDesignOperation}; we then describe our high-fidelity FEM simulation of its mechanics in Section~\ref{Subsec:SOFASim}.

\subsection{Soft Robot Design and Operation}
\label{subsec:SoftRobotDesignOperation}
\begin{figure}[t]
\centerline{\includegraphics[width=2.85in]{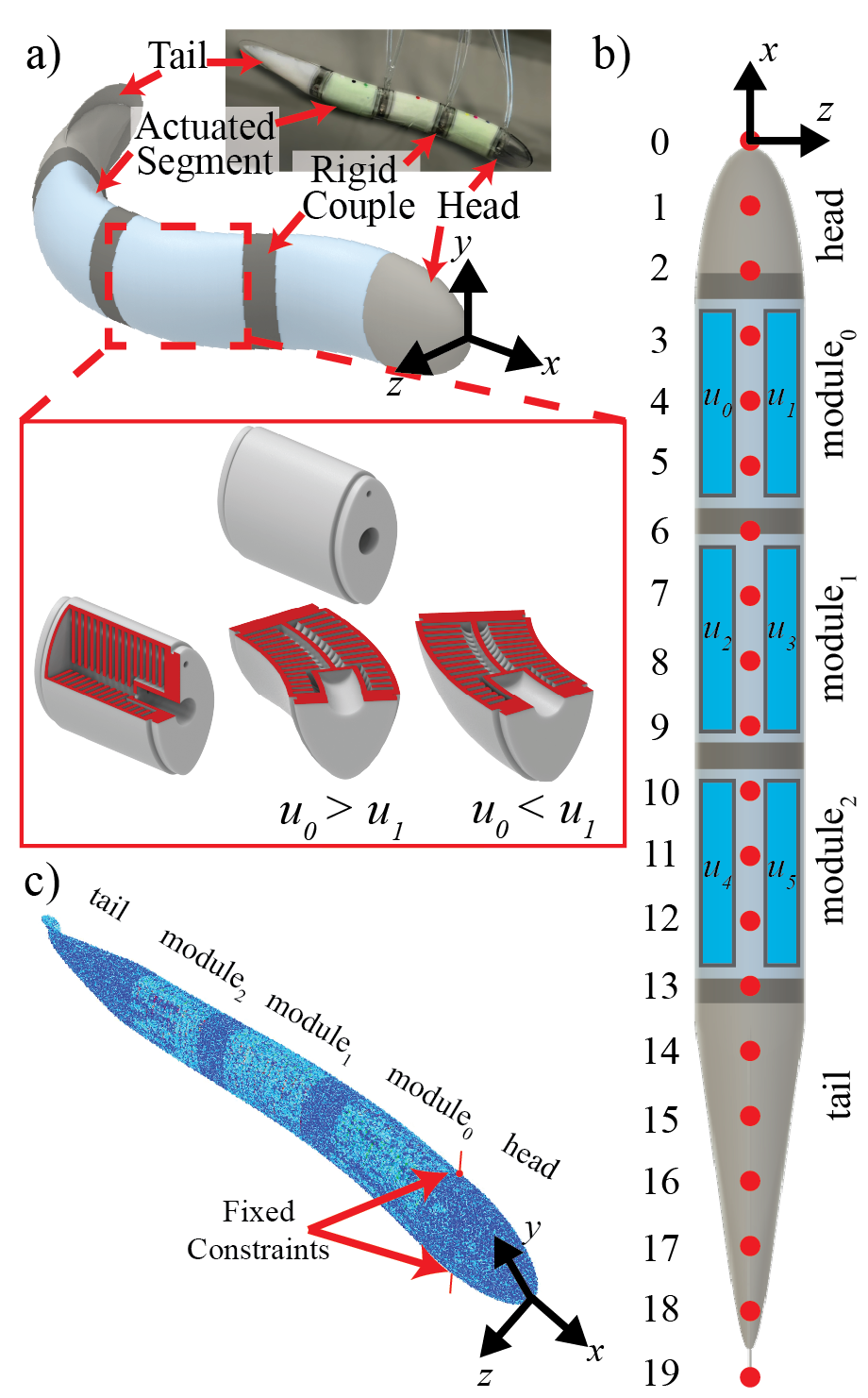}}
\caption{Anguilliform soft robot design and simulation. (a) We demonstrate our proposed methods on a simulated version of an eel-inspired soft robot developed in \cite{cervera-torralba_lost-core_2024,park_analysis_2025} (top right). This robot is comprised of soft, unactuated head and tail segments, three actuated segments driven by antagonistic fluid elastomer actuators, and rigid couples, used to securely connect the soft segments. As the two fluidic chambers in the actuated segments are pressurized with inputs, $\bu_i$, bending moments can be induced along the robot's body to affect its shape. b) The robot accepts six pressure inputs that are physically coupled such that $\bu_{2i} = -\bu_{2i+1}$ for $i\in \{0,1,2\}$. In our setting, we constructed the output of the robot as the $x$-$z$ position of 20 equally spaced points placed along the centerline of the robot's body, shown as red dots. c) We simulated the system's dynamics through a custom finite element model constructed in an open source simulation framework (SOFA), with high-dimensional state $\bx(t)\in\mathbb{R}^n$ representing the spatial coordinates of each node in a high-dimensional mesh of the robot's geometry.  We applied fixed constraints to two points at the base of the robot's head, allowing it to pivot about the line passing through these points (dorsal constrained point and line passing through both constrained points shown in red on the mesh of the robot). We selected this constraint based on past work on kinematics of anguilliform swimming indicating that this region of eels exhibits little lateral motion during forward swimming.}
\label{fig2}
\end{figure}

We considered the physical eel-inspired soft robot developed in \cite{cervera-torralba_lost-core_2024} and \cite{park_analysis_2025}. Our simulated testbench models the structural mechanics of an increased-scale version of this soft robot, with dimensions 1117~mm$\times$100~mm$\times$166.7~mm (L$\times$W$\times$H)---three times the size of the physical system~\cite{cervera-torralba_lost-core_2024}. This robot is comprised of five soft segments---passive soft head and tail segments and three actuated body segments---all coupled via custom rigid clamps (Fig.~\ref{fig2}.a). Each actuated body segment in the robot is driven by two antagonistic fluid elastomer actuators (FEAs) embedded inside of the body segment and separated by a thin, strain-limiting layer that is aligned with the sagittal plane of the robot \cite{cervera-torralba_lost-core_2024}. Thus, pressurizing the FEAs inside a given body segment induces a bending moment along the segment, and attaching multiple body segments together enables complex bending across the robot's body. 

In \cite{park_analysis_2025}, the antagonistic FEAs driving each body segment are pressurized by a gear pump in a closed hydraulic system. We approximate the behavior of this closed hydraulic drive system with a simulated constraint that when an FEA in a given body segment is actuated with pressure input, $u$, its antagonist FEA is actuated with the opposite pressure input, $-u$.
\subsection{Soft Robot Simulation in SOFA}
\label{Subsec:SOFASim}
Our numerical model builds on an open-source, finite element simulation framework (SOFA, SOFA Consortium)~\cite{payan_sofa_2012, coevoet_software_2017}. To ensure mesh compatibility with the simulation, we generated a mesh of the system using an open-source meshing tool (Gmsh~\cite{geuzaine_gmsh_2009}) that enabled multi-resolution meshing across the robot. During meshing, we enforced a finer mesh resolution in regions with more intricate features, such as the internal chamber geometry of the FEAs, and coarser resolution elsewhere. 

To avoid the computational expense and potential instability that can stem from resolving contact constraints, we meshed the robot as one contiguous body and defined regions of varying elastic moduli across the robot's body directly in SOFA. We used linear elasticity models across the entire system's structure, meaning that any nonlinear dynamic behavior would stem from geometric effects. We assigned three moduli to represent the three primary components of the robot: For regions containing rigid couples, we assigned an elastic modulus of 400~MPa to approximate rigid body behavior; for the strain-limiting layer of each body segment, we assigned an elastic modulus of 25~MPa, approximating that of the fiberglass composite (Garolite FR4) used in the physical system; and for the remainder of the robot, we assigned a modulus of 2~MPa---the 100\% strain modulus of the silicone polymer used to construct the robot's body (Dragonskin 10, Smooth-On). Finally, we applied two, fixed Lagrangian constraints at nodes located at the dorsal and ventral extremes near the base of the robot's head, allowing the system to pivot about the line connecting these two constraints (Fig~\ref{fig2}.c). Along with providing simulation stability, our placement of these boundary conditions was motivated by the fact that during straight-line locomotion, anguilliform swimmers tend to exhibit the least lateral motion in this region\cite{tytell_hydrodynamicsI_2004, tytell_hydrodynamicsII_2004}.

The resulting simulation models the nonlinear full-order dynamics of the robot
\begin{align}
\label{nonlinearFOM}
    \dot{\bx}(t)&= \bf{f}(\bx(t), \bu(t))\\
\label{nonlinearFOMOutput}
    \by(t)&= \bC \bx(t)
\end{align}
\noindent where at time $t$, the full-order state $\bx(t)\in\mathbb{R}^{n}$, represents the spatial coordinates of every node in the meshed geometry of the robot, the control input, $\bu(t)\in\mathbb{R}^{m}$, represents the pressure inputs to each of the fluidic chambers in the robot's actuators, and the output, $\by(t)\in\mathbb{R}^{p}$ represents the $x$-$z$ coordinates of 20 equally spaced points placed along the dorsal centerline of the robot (see red dots in Fig.~\ref{fig2}.b), computed by averaging the $x$-$z$ positions of the 20 mesh points nearest each control point's nominal position along the centerline. All spatial coordinates are expressed with respect to a stationary reference frame that is coincident with the anterior-most tip of the robot when in its neutral configuration~(Fig. \ref{fig2}). The resulting full-order model of the robot's structural dynamics had state of dimension ${n = 243{,}789
}$, output dimension ${p = 40}$, and input dimension ${m = 6}$, with the aforementioned antagonistic coupling constraints requiring only the selection of three decoupled control inputs at each timestep.

To numerically evaluate this model, we used an implicit, first-order integration scheme (\textsf{EulerImplicitSolver} component in SOFA) and parallelized linear solver (\textsf{ParallelCGLinearSolver} component in SOFA's \textsf{MultiThreading} plugin) with a fixed timestep of ${\Delta t = 0.01}$~seconds to maintain simulation stability. Due to the large size of the simulated robot's mesh, each simulation timestep required approximately 15~seconds of compute-time to evaluate on a custom desktop computer equipped with an Intel Core i9-10920X CPU, NVIDIA GeForce RTX 2080 Ti GPU, and 128.0~GiB of DDR4 memory, and running Ubuntu 22.04.5 LTS. 

\section{Dynamic Shape Control via Reduced-Order Model Predictive Control}
\label{sec:rompcFormulation}

Our numerical simulation provides a high-fidelity model of the anguilliform robot's structural dynamics. However, using this model in a real-time online control setting would be intractable, given the high-dimensionality of the full-order model. 

\begin{figure}[tb]
\centerline{\includegraphics[width=3.00in]{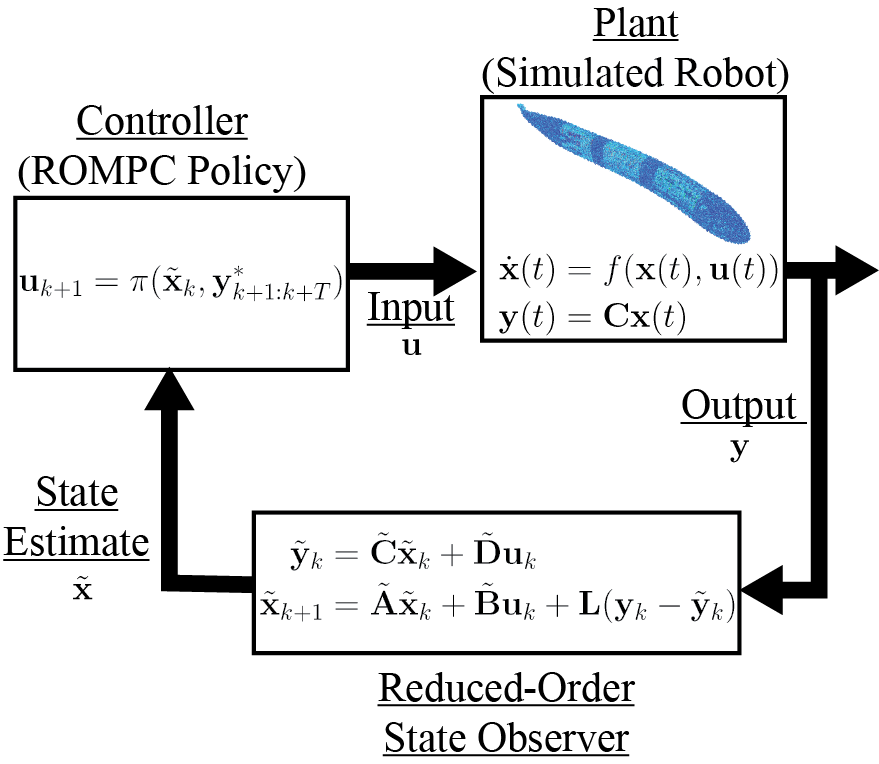}}
\caption{Block diagram of proposed reduced-order model predictive control loop. For each tested ROM, we synthesized a closed-loop state observer that takes measurements of the simulated robot's centerline to produce estimates of the reduced-order state. These estimates and a desired reference trajectory are then used to seed an optimization-based control policy, which produces a trajectory of input pressures. At each timestep, the pressures at the first timestep of this trajectory are used as input to the robot.}
\label{blockDiagram}
\end{figure}

To address this intractability, we propose a model-predictive control scheme that leverages non-intrusive ROMs produced from data generated by the full-order simulation, which we illustrate in Figure~\ref{blockDiagram}~\cite{lorenzetti_linear_2022, li_comprehensive_2022}. We focus on the use of linear ROMs as they are fasted and enable the use of mature machinery for high-speed and constraint-aware online control~\cite{borrelli_predictive_2017}. 

Consider a discrete-time, linear ROM of the form
\begin{align*}
    \tilde{\bx}_{k+1}&=\tilde{\bA}\tilde{\bx}_k + \tilde{\bB} \bu_k, \\
    \tilde{\by}_{k+1}&=\tilde{\bC} \tilde{\bx}_{k+1} + \tilde{\bD} \bu_{k+1},
\end{align*}

\noindent with reduced-order state $\tilde{\bx}_k\coloneq \tilde{\bx}(t_k)\in\mathbb{R}^r$ (with $r\ll n$) at time $t_k$, estimate of the full-order output $\tilde{\by}_{k}\coloneq \tilde{\by}(t_k)\in\mathbb{R}^p$,  and system matrices $\tilde{\bA}\in\mathbb{R}^{r\times r}$, $\tilde{\bB}\in\mathbb{R}^{r\times m}$, $\tilde{\bC}\in\mathbb{R}^{p\times r}$, and $\tilde{\bD}\in\mathbb{R}^{p\times m}$. While the full-order simulation generates outputs as a function of only the full-order state, we include the matrix $\tilde{\bD}$ here to accommodate reduced-order modeling methods that produce a feedthrough term, such as the ERA method described in Section~\ref{sec: era}.

We use a Luenberger state observer that produces reduced-order state estimates to seed an optimization-based control policy (Fig.~\ref{blockDiagram}).  
At each timestep $k$, the centerline output $\by_k\coloneq \by(t_k)$ from the nonlinear FOM \eqref{nonlinearFOM}-\eqref{nonlinearFOMOutput} is used to update the observer's reduced-order state estimate $\tilde{\bx}_k$ under the estimation law
\begin{subequations}
\label{eq:estLaw_CL}
\begin{align}
    \label{subeq:estLaw_CL_outputEst}
    \tilde{\by}_k &= \tilde{\bC}\tilde{\bx}_{k} +  \tilde{\bD}\bu_{k}\\
    \label{subeq:estLaw_CL_stateUpdate}
    \tilde{\bx}_{{k}} &= \tilde{\bA}\tilde{\bx}_{k-1} +  \tilde{\bB}\bu_{k-1} + \bL(\by_k-\tilde{\by}_k )
\end{align}
\end{subequations}

The updated state estimate produced in~\eqref{subeq:estLaw_CL_stateUpdate} is then used as the initial condition of the policy $\pi(\tilde{\bx}_k,\by^*_{k+1:k+T})$, that, given the reference trajectory $\by^*_{k+1:k+T}$, over control horizon $T$, computes an optimal control sequence $\bu^*_{k+1:k+T}$, by solving the quadratic program at every timestep $k$,
\begin{subequations}
\label{eq:rompc}
\begin{align}
\begin{split}
\label{subeq:rompc_obj}
    \bu^*_{k+1:k+T} &= \underset{\bu_{k+1:k+T}}{\arg\min}\sum_{i=k+1}^{k+T} \Bigl( \lVert\tilde{\by}_i - \by^*_{i}\rVert_{\bW_\by}^2 + \lVert\bu_i\rVert_{\bW_\bu}^2 \Bigr. \\
    & \Bigl. \qquad \qquad \qquad \qquad \qquad + \lVert\bu_i-\bu_{i-1}\rVert_{\bW_{\bdu}}^2 \Bigr) 
\end{split}\\
\text{s.t.} \qquad & \nonumber\\
\label{subeq:rompc_constraint_initState}
\tilde{\bx}_{-1} &=\tilde{\bx}_{k}\\
\label{subeq:rompc_constraint_initInput}
\bu_{-1} &=\bu_{k}\\
\label{subeq:rompc_constraint_state}
\tilde{\bx}_{i} &= \tilde{\bA}\tilde{\bx}_{i-1} + \tilde{\bB}\bu_{i-1}, \\
\label{subeq:rompc_constraint_output}
\tilde{\by}_i &= \tilde{\bC}\tilde{\bx}_i + \tilde{\bD}\bu_i, \\
\label{subeq:rompc_constraint_maxInput}
\lVert \bu_{i}\rVert_\infty &< u_{\max}, \\
\label{subeq:rompc_constraint_inputAntag}
\bu_{i,2j}  &= -\bu_{i,2j+1}
\qquad \forall j \in \{0, 1, 2\} 
\end{align}
\end{subequations}
\noindent where $\lVert\cdot\rVert_{\bW_*}$ indicates the weighted 2-norm of a vector with diagonal weighting matrix $\bW_*$. The objective of this optimization is to determine the sequence of inputs that simultaneously minimizes the weighted magnitudes of the tracking error $\lVert\tilde{\by}_i - \by^*_{i}\rVert_{\bW_\by}^2$,  the weighted control energy $\lVert\bu_i\rVert_{\bW_\bu}^2$, and the discrete rate of change of the control input $\lVert\bu_i-\bu_{i-1}\rVert_{\bW_{\bdu}}^2$. The constraints in \eqref{subeq:rompc_constraint_initState} and \eqref{subeq:rompc_constraint_initInput} are used to set the initial state and input for the optimization, where the initial state constraint uses the estimate in \eqref{subeq:estLaw_CL_stateUpdate}. These initial conditions are then rolled out over the control horizon via constraint \eqref{subeq:rompc_constraint_state}, with outputs evaluated at each timestep through constraint \eqref{subeq:rompc_constraint_output}. Finally, the constraint in \eqref{subeq:rompc_constraint_maxInput} is used to set a maximum possible pressure value for control and the constraint in \eqref{subeq:rompc_constraint_inputAntag} is used to account for the antagonistic coupling of pressures present in the robot's closed fluidic drive system. After computing the optimizing input sequence $\bu^{*}_{k+1:k+T}$, from this constrained optimization, we apply the first input in the sequence, $\bu^*_{k+1}$, to the simulated robot, compute the system's response to the optimized control input, and generate the simulated systems state at the next timestep. We continue iterating this process until the end of the simulation horizon.

\section{Data-Driven Reduced-Order Modeling}
\label{sec:DataDrivenROMs}
For the controlled simulated system to effectively track full-body reference trajectories, the linear ROM used for state estimation and control optimization must effectively capture the controlled dynamics of the robot. Thus, we compared multiple methods for data-driven, linear model reduction with control-affine inputs with the goal of assessing their performance in a dynamic, full-body control setting.

In our comparative study, we considered three methods representing different classes of data-driven model reduction/system identification---the Lagrangian Operator Inference method (LOpInf) \cite{sharma2024preserving}, dynamic mode decomposition with control (DMDc) \cite{proctor_dynamic_2016}, and a combination of the Eigensystem realization algorithm (ERA) with the observer Kalman filter identification algorithm (OKID) \cite{juang_identification_1993, juang_eigensystem_1985}. In this section, we describe each method as used in our setting, beginning with details how we preprocessed data for model synthesis in Section~\ref{subsec:DataPreprocessing}, and then introducing formulations for the LOpInf, DMDc, and ERA methods in Section~\ref{sec:lopinf}, Section~\ref{sec:dmdc}, and Section~\ref{sec: era} respectively.

\subsection{Data Preprocessing}
\label{subsec:DataPreprocessing}
We assume access to the full internal state, $\bx_k\in \real^n$, control inputs that generated that state $\bu_k \in \real^m$,  and resulting output $\by_k\in \real^p$, where the subscript $k$ indicates the value of a variable at time $t_k$. Since our main goal was to generate data-driven ROMs for controlling the simulated soft robot, we pre-processed the full-order states and outputs of all simulation data to compute centered data: to compute centered full-order state and output trajectories $\bx_{\text{centered},k} = \bx_k-\bx_{\text{neutral}}$ and $\by_{\text{centered},k}=\by_k-\by_{\text{neutral}}$, respectively. These centered variables are shifted such that nonzero coordinates of the neutral state $\bx_{\text{neutral}}\in \mathbb{R}^n$ and output $\by_{\text{neutral}}\in~ \real^p$ of the robot map to the origin of the full-order state and output spaces, respectively. This centering of state and output trajectories was motivated by the system-theoretic insight that the only potential stable point of a linear system subject to zero input and zero disturbance is at its origin, and that this should coincide with the stable configuration of the robot when subject to zero input and zero disturbance. 
We synthesized all system models using these centered variables, un-centering them only for the sake of mesh visualization. From hereon, we simplify the notation for the centered variables: the centered state trajectory $\bx_{\text{centered},k}$ and the centered output trajectory $\by_{\text{centered},k}$ at time $t_k$ are denoted as $\bx_k$ and $\by_k$, respectively.

\subsection{Lagrangian Operator Inference Method}
\label{sec:lopinf}
The Lagrangian Operator Inference (LOpInf) method is a non-intrusive model reduction method that first projects high-dimensional snapshot data onto a low-dimensional subspace, and then fits linear reduced operators that preserve an assumed Lagrangian structure of the system's dynamics~\cite{sharma2024preserving}. To learn a ROM with this method, we first construct snapshot matrices of the full-order state, control input, and output, as 
\begin{align}
    \bX&=[\bx_1, \cdots, \bx_{K}] \in \real^{n \times K},
    \label{eq:snapshot_LOpInf} \\
    \bU&=[\bu_1,\cdots,\bu_{K}] \in \real^{m \times K}, 
    \label{eq:snapshot_input} \\ 
\bY&=[\by_1, \cdots, \by_{K}] \in \real^{p \times K},
      \label{eq:snapshot_output}
\end{align}
\noindent where $K$ is the number of snapshots.

This data is very high-dimensional ($n = 243{,}789
$), and we cannot learn from it directly. Thus, to project the high-dimensional snapshot data onto a low-dimensional subspace, we first construct a basis matrix $\bcU_r$ via truncated singular value decomposition of $\bX$ with truncation value $r$ as

\begin{equation}
        \bX\approx \bcU_r\pmb\Sigma_r\bcV_r^*,
\end{equation}
where $\bcU_r \in \real^{n \times r}$, $\pmb \Sigma_r \in \real^{r \times r}$, and $\bcV_r^* \in \real^{r \times K}$.

We then construct a reduced snapshot matrix by projecting the full-order snapshot data onto the reduced basis as
\begin{equation}
    \widehat{\bX}=\bcU_r^\top\bX=[\widehat{\bx}_1, \cdots, \widehat{\bx}_{K}] \in \real^{r \times K},
    \label{eq:reduced_statedata}
\end{equation}
where $\widehat{\bx}_k:=\bcU_r^\top\bx_k$ is the reduced state trajectory at time $t_k$. We also build snapshot matrices of the reduced first-order and second-order time-derivative data as
\begin{equation*}
    \begin{aligned}
      \dot{\widehat{\bX}}=[\dot{\widehat{\bx}}_1, \cdots, \dot{\widehat{\bx}}_{K}] \in \real^{r \times K}, \quad 
      \ddot{\widehat{\bX}}=[\ddot{\widehat{\bx}}_1, \cdots, \ddot{\widehat{\bx}}_{K}] \in \real^{r \times K},
    \end{aligned}
\end{equation*}
    where $\dot{\widehat{\bx}}_k$ and $\ddot{\widehat{\bx}}_k$ are obtained from the reduced state trajectories via numerical approximation, here with an eighth-order central finite difference scheme. 
Given the snapshot data matrices of the state and its derivatives, we postulate a Lagrangian formulation for the ROM:
    \begin{align}
    \label{eq:L_rom}
    \ddot{\widehat{\bx}}(t) + \widehat{\mathbf D}\dot{\widehat{\bx}}(t) + \hbK\widehat{\bx}(t)=\hbB \bu(t),\\
    \by(t)=\hbC \widehat{\bx}(t),
    \end{align}
with the reduced damping matrix $\widehat{\mathbf D} \in \real^{r \times r}$, the reduced stiffness matrix $\hbK \in \real^{r \times r}$, the reduced input matrix $\hbB \in \real^{r \times m}$, and the reduced output matrix $\hbC \in \real^{p \times r}$.

Given the reduced snapshot data $\widehat{\bX}$ and the reduced time-derivative data $\dot{\widehat{\bX}}$ and $\ddot{\widehat{\bX}}$, we infer the Lagrangian ROM operators in~\eqref{eq:L_rom} by solving
\begin{equation}
\min _{\substack{\hbK=\hbK^\top \succ 0, \widehat{\mathbf D}=\widehat{\mathbf D}^\top \succ 0, \hbB}}
\lVert \ddot{\widehat{\bX}} + \widehat{\mathbf D}\dot{\widehat{\bX}} + \hbK\widehat{\bX}- \hbB \bU \rVert_{F}.
\label{eq:lopinf}
\end{equation}
We infer the reduced output operator $\hbC$ by solving the least-squares problem 
\begin{equation}
    \min _{\hbC}
\lVert \bY - \hbC\hbX \rVert_{F}.
\label{eq:lopinf_output}
\end{equation}
The symmetric positive-definite constraints on $\hbK$ and $\widehat{\mathbf D}$ in~\eqref{eq:lopinf} ensure that the linear ROM~\eqref{eq:L_rom} preserves the underlying Lagrangian structure (see \cite{sharma2024preserving} for details).

The resulting ROM defines second-order, continuous-time dynamics that, for the sake of consistency in our MPC formulation, we recast as a system of first-order, discrete-time difference equations. To do this, we first define an augmented reduced state, $\tilde{\bx}^{(e)} = \left[\widehat{\bx}^\top,\dot{\widehat{\bx}} ^\top\right]^\top$, with equivalent continuous-time dynamics
\begin{align}
    \dot{\tilde{\bx}}^{(e)}(t) &= \widehat{\bA}\tilde{\bx}^{(e)}(t) + \widehat{\bB}\bu(t),\\
    \by(t) &= \widehat{\bC}\tilde{\bx}^{(e)}(t),
\end{align}
\noindent where, with a slight abuse of notation, we have
\begin{equation*}
        \widehat{\bA} = \begin{bmatrix}
        \bzero_{r\times r} & \Ir\\
        -\hbK & -\widehat{\mathbf D}
    \end{bmatrix}
    ,\quad 
    \widehat{\bB}=\begin{bmatrix}
        \bzero_{r\times m}\\
        -\hbB
    \end{bmatrix}, \quad  
    \widehat{\bC} = \begin{bmatrix}
        \hbC &
        \bzero_{p\times r}
    \end{bmatrix}.
\end{equation*}

Assuming zero-order hold conditions on the input, $\bu_k$, and sampling time $T_s = \Delta t = 0.01$~seconds, we derive an equivalent discrete-time system \cite{hespanha_linear_2018} as 
\begin{align}
    \tilde{\bx}^{(e)}_{k+1} &= \tilde{\bA} \tilde{\bx}^{(e)}_{k} + \tilde{\bB}\bu_k,\\
    \tilde{\by}_k &= \tilde{\bC}\tilde{\bx}^{(e)}_{k},
\end{align}
with $
    \tilde{\bA} = e^{\widehat{\bA}T_s}$,
    $\tilde{\bB}= \widehat{\bA}^{-1}(\tilde{\bA}-\Ir)\widehat{\bB}$, and
    $\tilde{\bC} =\widehat{\bC}$.

\subsection{Dynamic mode decomposition with control}
\label{sec:dmdc}
Dynamic mode decomposition with control (DMDc)~\cite{proctor_dynamic_2016} is a non-intrusive model reduction method that learns linear ROMs of high-dimensional nonautonomous systems purely from data. Similar to DMD \cite{schmid2010dynamic, rowley2009spectral}, we first build a snapshot data matrix $\bX=[\bx_1, \cdots, \bx_{K-1}] \in \real^{n \times K-1}$, and a time-shifted snapshot data matrix $\bX'=[\bx_2, \cdots, \bx_{K}] \in \real^{n \times K-1}$. We also build the corresponding input snapshot matrix, $\bU=[\bu_1,\cdots, \bu_{K-1}]\in \real^{m \times K-1}$, and stack data matrices $\bX$ and $\bU$ to construct $\pmb{\Omega}:=\begin{bmatrix} \bX \\ \bU  \end{bmatrix} \in \real^{n+m \times K-1}$. 

\vspace{1mm}
Given data matrices $\pmb{\Omega}$ and $\bX'$, we compute the truncated singular value decompositions of $\pmb{\Omega}$ and $\bX'$ with truncation values of $q$ and $r$, respectively, resulting in the decompositions
\begin{equation}
\pmb{\Omega}\approx \begin{bmatrix} \tilde{\bcU}_1 \\ \tilde{\bcU}_2  \end{bmatrix}\tilde{\pmb\Sigma}\tilde{\bcV}^*
\qquad
\bX'\approx \widehat{\bcU}\widehat{\pmb \Sigma}\widehat{\bcV}^*,
\end{equation}
with $\tilde{\bcU}_1 \in \real^{n \times q}$, $\tilde{\bcU}_2 \in \real^{m \times q}$, $\tilde{\pmb \Sigma} \in \real^{q \times q}$, $\tilde{\bcV}^*\in \real^{q \times K-1}$, $ \widehat{\bcU} \in \real^{n \times r}$, $\widehat{\pmb \Sigma} \in \real^{r \times r}$, and $\widehat{\bcV}^* \in \real^{r \times K-1}$. Using the full-order state approximation $\bx\approx\widehat{\bcU}\, \hbx$, the resulting DMDc linear ROM is
\begin{align}
\tilde{\bx}_{k+1}&=\tilde{\bA} \, \tilde{\bx}_k + \tilde{\bB} \bu_k, \\
\tilde{\by}_{k+1}&=\tilde{\bC} \tilde{\bx}_k,
\end{align}
with the reduced state matrix $\tbA:=\widehat{\bcU}^*\bX'\tilde{\bcV}\tilde{\pmb\Sigma}^{-1}\tilde{\bcU}_1^*\widehat{\bcU} \in \real^{r \times r}$, the reduced input matrix $\tbB:=\widehat{\bcU}^*\bX'\tilde{\bcV}\tilde{\pmb\Sigma}^{-1}\tilde{\bcU}_2^* \in \real^{m \times r}$, and the reduced output matrix $\tilde{\bC}:=\widehat{\bC}\widehat{\bcU}$ where  
\begin{equation}
    \widehat{\bC}=\min _{\widehat{\bC}}
\lVert \bY - \widehat{\bC}\,\bX \rVert_{F},
\label{eq:dmdc_output}
\end{equation}
where $\bY$ is as defined in \eqref{eq:snapshot_output}.
\subsection{Eigensystem Realization Algorithm}
\label{sec: era}
The Eigensystem Realization Algorithm (ERA)~\cite{juang_eigensystem_1985} is a system identification method for identifying low-dimensional linear state-space models from impulse response data. For arbitrary input-output data, the ERA approach is used in combination with the Observer Kalman Filter Identification (OKID) algorithm~\cite{juang_identification_1993}, which estimates impulse response data from arbitrary input-output data. To employ this method, we first construct the output data matrix $\bY$, as defined in \eqref{eq:snapshot_output}, as well as the input data matrix
\begin{equation}
    \pmb{\mathfrak{U}} =\begin{bmatrix}
        \bu_1 & \bu_2 &\bu_3&\cdots &\bu_{K}\\
        0 &\bu_1 &\bu_2 &\cdots &\bu_{K-1}\\
        &&&\ddots&\\
        0 & 0 &0 &\cdots &\bu_1\\
        
    \end{bmatrix}\in \mathbb{R}^{mK \times K}
\end{equation}
using the input and output data collected from a single experiment. The input-output relationship of a linear system with arbitrary input can be expressed as 
\begin{equation}
    \bY = \bM \pmb{\mathfrak{U}}, 
\end{equation}
where 
\begin{equation}
    \bM = \begin{bmatrix}
        \bD&\bC\bB&\bC\bA\bB&\cdots &\bC\bA^{K-1}\bB
    \end{bmatrix}
\end{equation}
is a matrix of Markov parameters for the system, written in terms of the unknown $\bA$, $\bB$, $\bC$ and input-output feedthrough matrix, $\bD$. Recognizing this relationship, we approximate $\bM$ as
\begin{equation}
    \widehat{\bM} = \bY \pmb{\mathfrak{U}}^\dagger, 
\end{equation} 
where $\pmb{\mathfrak{U}}^\dagger$ denotes the Moore-Penrose pseudo-inverse of the input matrix, $\pmb{\mathfrak{U}}$~\cite{penrose1955generalized}. Using the estimated Markov parameters in $\widehat{\bM}$, we then construct two Hankel matrices 
\begin{equation}
    \bH = \begin{bmatrix}
        \bC\bB          & \bC\bA\bB       & \cdots & \bC\bA^{T_H-1}\bB\\
        \bC\bA\bB         & \bC\bA^2\bB     & \cdots & \bC\bA^{T_H}\bB  \\
        \vdots      &           & \ddots &           \\
        \bC\bA^{T_H-1}\bB & \bC\bA^{T_H}\bB & \cdots & \bC\bA^{2T_H-1}\bB  
    \end{bmatrix}
\end{equation}
and
\begin{equation}
    \bH' = \begin{bmatrix}
        \bC\bA\bB       & \bC\bA^2\bB       & \cdots & \bC\bA^{T_H}\bB\\
        \bC\bA^2\bB     & \bC\bA^3\bB       & \cdots & \bC\bA^{T_H+1}\bB\\
        \vdots    &             & \ddots &          \\
        \bC\bA^{T_H}\bB & \bC\bA^{T_H+1}\bB & \cdots & \bC\bA^{2T_H}\bB  
    \end{bmatrix},
\end{equation}
where $T_H = \lfloor K/2\rfloor-1$.
After taking an order-$r$ truncated singular value decomposition of $\bH$, 
\begin{equation}
    \bH \approx \tilde{\bcU}_r \tilde{\pmb\Sigma}_r \tilde{\bcV}_r^\top, 
\end{equation}
we estimate the order-$r$ system matrices as 
\begin{align}
    \tilde{\bA} &=  \tilde{\pmb \Sigma}_r^{-1/2}\tilde{\bcU}_r^\top\bH'\tilde{\bcV}_r\tilde{\pmb \Sigma}_r^{-1/2}, \\
    \tilde{\bB} &= \tilde{\pmb\Sigma}_r^{1/2}\tilde{\bcV^\top}_r \bE_l, \\
    \tilde{\bC} &= \bE_m^\top \tilde{\bcU}_r\tilde{\pmb\Sigma}_r^{1/2},
\end{align}
and $\tilde{\bD}$ is taken directly from the first block entry of the Markov parameter matrix $\bM$. The matrices $\bE_l$ and $\bE_m$ are selection matrices constructed as
\begin{equation*}
    \bE_l=~[\mathbf I_{l},\bzero_{l\times (l(T_H-1))} ]^\top, \; 
    \bE_m=~[\mathbf I_{m},\bzero_{m\times (m(T_H-1))}]^\top.
\end{equation*}

The result of this procedure is a linear, order-$r$ ROM of the form
\begin{align*}
    \tilde{\bx}_{k+1}&=\tilde{\bA}\tilde{\bx}_k + \tilde{\bB} \bu_k, \\
    \tilde{\by}_{k+1}&=\tilde{\bC} \tilde{\bx}_{k+1} + \tilde{\bD} \bu_{k+1},
\end{align*}
with reduced state matrix $\tbA \in \real^{r \times r}$, reduced input matrix $\tbB \in \real^{r \times m}$, $\tbC \in \real^{p \times r}$, and feedforward matrix $\tbD \in \real^{p \times m}$. Unlike LOpInf and DMDc, the ERA/OKID approach is based entirely on the input-output data and does not require a snapshot data matrix containing full-order state. This can be beneficial in settings where full-order state data is not available.

\section{ROM Training and Estimation Accuracy}
\label{sec:romTraining}
Using the methods presented in the previous section, we generated a total of 70~ROMs and assessed their predictive accuracy as a function of their dimension and the amount of data used for training. In this section we detail this procedure, beginning with how we generated a large-scale dataset of the simulated robot's dynamics in Section~\ref{subsec:DatasetGeneration}, then in Section~\ref{subsec:ROMTraining} defining how we generated ROMs from each method with varying dimensions and amounts of training data. After, we introduce the estimation laws that we used to assess the predictive accuracy of each ROM in Section~\ref{sec:EstimationLaws}, and characterize the predictive performance of each ROMs under these estimation laws in Section~\ref{subsec:EffectsOfTrainingHyperparametersOnEstimationAccuracy}.
\subsection{Dataset Generation}
\label{subsec:DatasetGeneration}
We constructed a dataset of ${N=40}$ simulated trials of the simulated soft robot, with each trial comprised of ${T=1000}$ timesteps, resulting in full-order state, input, and output trajectories that each span 10~seconds. For all trials, we prescribed sinusoidal pressure inputs to the channels in the soft actuators shown in Fig.~\ref{fig2} so as to excite cyclic behavior in the robot's dynamics. These inputs are defined as
\begin{subequations}
    \begin{align}
    u_0(t) &= A_1\sin(2\pi f t + \phi_0), \qquad u_1(t)= -u_0(t), \\
    u_2(t) &= A_2\sin(2\pi f t + \phi_1), \qquad u_3(t)= -u_2(t), \\
    u_4(t) &= A_3\sin(2\pi f t + \phi_2), \qquad u_5(t)= -u_4(t),
    \end{align}
\end{subequations}
where $A_1, A_2,$ and $A_3$ denote the amplitudes of the sinusoidal pressure inputs to each segments, $f$, is the frequency of oscillation, and ${\phi_1=0^\circ}$, ${\phi_2=120^\circ}$, and ${\phi_3=240^\circ}$ are the relative phases between inputs to each segment. As mentioned above, $u_{2i} = -u_{2i+1}, i\in\{0,1,2\}$ to approximate the physical constraints imposed by the closed fluidic system used to drive each actuated body segment antagonistically \cite{park_analysis_2025}. 

To excite a variety of dynamic behaviors, we simulated the anguilliform robot over $N=40$ trials using five different frequencies and eight combinations of amplitudes. We chose frequencies that spanned from an approximately quasi-static regime (${f=0.1}$~Hz) to frequencies seen in anguilliform animals during dynamic swimming (${f=1.5}$~Hz) \cite{tytell_hydrodynamicsII_2004}. Additionally, we chose amplitudes with the goal of exciting varying degrees of geometric nonlinearity. For this, we selected both low amplitude, $A_L$, and high amplitude, $A_H$, inputs to apply to active segments in a trial. We selected $A_H$ such that the steady-state  bending it induces in a given actuated segment is similar to that demonstrated by the physical platform in~\cite{park_analysis_2025}. A full description of actuation parameters for generating the dataset is shown in Table~\ref{tab:DatasetParameters}.

During each trial, we collected 1) the inputs signals applied to the robot; 2) the full-order state of the robot; and 3) the output of the system, defined by the $x-z$ position of 20 equally spaced points along the dorsal centerline of the robot's body.

\subsection{ROM Training}
\label{subsec:ROMTraining}
Once we select the ROM model class, the fidelity of the data-driven ROMs generated by the ERA, DMDc, and LOpInf methods are defined by two factors: 1) the amount and quality of data used during training, and 2) the dimension $r$, of the ROM being synthesized. 
To understand the effects of these factors on the estimation and control of a soft robot, we used each method to synthesize ROMs of dimension $r=2,4, ..., 20$ on datasets that included one, two, and three trials of simulated data. Due to the high dimensionality of the full-order system data, three trials proved to be the limit for local training on the custom desktop computer described in Section~\ref{Subsec:SOFASim} and the use of an off-the-shelf parallel computation library (Dask, Dask Core). Due to the formulation of the ERA method, models could only be synthesized on one trial of timeseries data. Thus, our experimental study considered 70 different ROMs---30 each generated with DMDc and LOpInf, and 10 generated with ERA.

We reserved three trials for training each ROM (highlighted in green in Table \ref{tab:DatasetParameters}) based on the heuristic that the input parameters for these trials would produce a diverse set of dynamic and quasi-static behaviors that could then be represented by the resulting ROMs. We used the data generated by Trial~40 for all ROMs trained on a single trial of data, assuming the high-frequency and high-amplitude actuation would excite the richest set of structural dynamic behaviors---all ERA-based ROMs were trained on only this trial. We used Trials~40 and 17 for all ROMs trained on two trials of data, and additionally included Trial~36 for all ROMs trained on 3 trials of data. These trials provided additional information on lower-frequency dynamics with varying amplitudes. As geometric effects were the primary source of nonlinearity in the simulated system, we included trials with varied input amplitude to introduce varying amounts of nonlinearity into training.  
    
\begin{table}[tb]
\begin{center}
\caption{Open-Loop Actuation Parameters for Dataset Generation. Trials reserved for model training are marked in green and trials used in control experiments are marked with $^*$.}
\label{tab:DatasetParameters}
\resizebox{3.25in}{!}{%
\renewcommand{\arraystretch}{1.7}
\begin{tabular}{cc|ccccc|}
\cline{3-7}
\multicolumn{1}{l}{}                                                                                            & \multicolumn{1}{l|}{}                              & \multicolumn{5}{c|}{\cellcolor[HTML]{C0C0C0}\textbf{Actuation Frequency, $f$ {[}Hz{]}}}                                                                                                                                                                                            \\ \cline{3-7} 
\multicolumn{1}{l}{}                                                                                            & \multicolumn{1}{l|}{}                              & \multicolumn{1}{c|}{\cellcolor[HTML]{EFEFEF}\textbf{0.1}} & \multicolumn{1}{c|}{\cellcolor[HTML]{EFEFEF}\textbf{0.3}} & \multicolumn{1}{c|}{\cellcolor[HTML]{EFEFEF}\textbf{0.5}} & \multicolumn{1}{c|}{\cellcolor[HTML]{EFEFEF}\textbf{1}} & \cellcolor[HTML]{EFEFEF}\textbf{1.5} \\ \hline
\multicolumn{1}{|c|}{\cellcolor[HTML]{C0C0C0}}                                                                  & \cellcolor[HTML]{EFEFEF}\textbf{$(A_L, 0, 0)$}     & \multicolumn{1}{c|}{1}                                    & \multicolumn{1}{c|}{2}                                    & \multicolumn{1}{c|}{3}                                    & \multicolumn{1}{c|}{4\textbf{*}}                                  & 5                                    \\ \cline{2-7} 
\multicolumn{1}{|c|}{\cellcolor[HTML]{C0C0C0}}                                                                  & \cellcolor[HTML]{EFEFEF}\textbf{$(0, A_L, 0)$}     & \multicolumn{1}{c|}{6\textbf{*}}                                    & \multicolumn{1}{c|}{7}                                    & \multicolumn{1}{c|}{8}                                    & \multicolumn{1}{c|}{9}                                  & 10                                   \\ \cline{2-7} 
\multicolumn{1}{|c|}{\cellcolor[HTML]{C0C0C0}}                                                                  & \cellcolor[HTML]{EFEFEF}\textbf{$(0, 0, A_L)$}     & \multicolumn{1}{c|}{11}                                   & \multicolumn{1}{c|}{12\textbf{*}}                                   & \multicolumn{1}{c|}{13}                                   & \multicolumn{1}{c|}{14}                                 & 15                                   \\ \cline{2-7} 
\multicolumn{1}{|c|}{\cellcolor[HTML]{C0C0C0}}                                                                  & \cellcolor[HTML]{EFEFEF}\textbf{$(A_L, A_L, A_L)$} & \multicolumn{1}{c|}{16}                                   & \multicolumn{1}{c|}{\cellcolor{green!20} 17}                                   & \multicolumn{1}{c|}{18}                                   & \multicolumn{1}{c|}{19\textbf{*}}                                 & 20                                   \\ \cline{2-7} 
\multicolumn{1}{|c|}{\cellcolor[HTML]{C0C0C0}}                                                                  & \cellcolor[HTML]{EFEFEF}\textbf{$(A_H, 0, 0)$}     & \multicolumn{1}{c|}{21}                                   & \multicolumn{1}{c|}{22\textbf{*}}                                   & \multicolumn{1}{c|}{23}                                   & \multicolumn{1}{c|}{24}                                 & 25                                   \\ \cline{2-7} 
\multicolumn{1}{|c|}{\cellcolor[HTML]{C0C0C0}}                                                                  & \cellcolor[HTML]{EFEFEF}\textbf{$(0, A_H, 0)$}     & \multicolumn{1}{c|}{26\textbf{*}}                                   & \multicolumn{1}{c|}{27}                                   & \multicolumn{1}{c|}{28}                                   & \multicolumn{1}{c|}{29}                                 & 30                                   \\ \cline{2-7} 
\multicolumn{1}{|c|}{\cellcolor[HTML]{C0C0C0}}                                                                  & \cellcolor[HTML]{EFEFEF}\textbf{$(0, 0, A_H)$}     & \multicolumn{1}{c|}{31}                                   & \multicolumn{1}{c|}{32}                                   & \multicolumn{1}{c|}{33\textbf{*}}                                   & \multicolumn{1}{c|}{34}                                 & 35                                   \\ \cline{2-7} 
\multicolumn{1}{|c|}{\multirow{-8}{*}{\rotatebox[origin=c]{90}{\cellcolor[HTML]{C0C0C0}\textbf{Amplitudes $(A_1, A_2, A_3)$}}}} & \cellcolor[HTML]{EFEFEF}\textbf{$(A_H, A_H, A_H)$} & \multicolumn{1}{c|}{\cellcolor{green!20} 36}                                   & \multicolumn{1}{c|}{37}                                   & \multicolumn{1}{c|}{38}                                   & \multicolumn{1}{c|}{39\textbf{*}}                                 & \cellcolor{green!20} 40                                   \\ \hline
\end{tabular}%
}
\end{center}
\vspace{-15pt}

\end{table}

\subsection{Open-loop and Closed-loop Estimation Laws}
\label{sec:EstimationLaws}
Our proposed ROMPC scheme (Fig.~\ref{blockDiagram}) leverages ROMs in both open-loop and closed-loop estimation settings described in~\cite{hespanha_linear_2018}, wherein ``open-loop" estimation refers to using a given initial reduced-order state and input sequence with the system matrices of a given ROM to estimate the reduced-order state and output trajectories of the simulated system. In contrast, ``closed-loop" estimation refers to additionally using output measurements at each timestep to correct the state estimates produced by the state estimator. As our optimization-based control policy requires rolling out trajectories to account for the constraints in \eqref{eq:rompc}, it is necessary for a given ROM to provide accurate open-loop predictions of the output trajectories over the time horizon of control. These roll outs must also be seeded with accurate estimates of the reduced-order state at a given timestep, provided by a closed-loop state observer~\cite{hespanha_linear_2018}. 

To assess the open-loop prediction accuracy of each ROM, we generated output estimates $\tilde{\by}_{k}$ using a chosen timeseries of inputs $\bu_k$, and initial condition $\tilde{\bx}_0 = \bzero$ with the estimation law
\begin{equation}
\label{eq:estLaw_OL}
\begin{aligned}
\tilde{\bx}_{{k+1}} &= \tilde{\bA}\tilde{\bx}_{k} +  \tilde{\bB}\bu_{k}
\\
\tilde{\by}_{k+1} &= \tilde{\bC}\tilde{\bx}_{k+1} +  \tilde{\bD}\bu_{k+1}.
\end{aligned}
\end{equation}
To assess the closed-loop estimation accuracy, we constructed a Luenberger observer defined in~\eqref{eq:estLaw_CL} with initial conditions $\tilde{\bx}_{0} = \bzero$ and $\tilde{\by}_{0} = \bzero$, and selecting the gain matrix $\bL$ for each ROM type and each dimension $r$, such that the $r$ eigenvalues of $\tilde{\bA}-\bL\tilde{\bC}$ were real and evenly distributed between -0.5 and 0.5. The values of $\by_{k}$ in~\eqref{eq:estLaw_CL} were taken from the output data of each trial. 

\subsection{Effects of Training Hyperparameters on Estimation Accuracy}
\label{subsec:EffectsOfTrainingHyperparametersOnEstimationAccuracy}
\begin{figure*}[htbp]
\centerline{\includegraphics[width=6.5in]{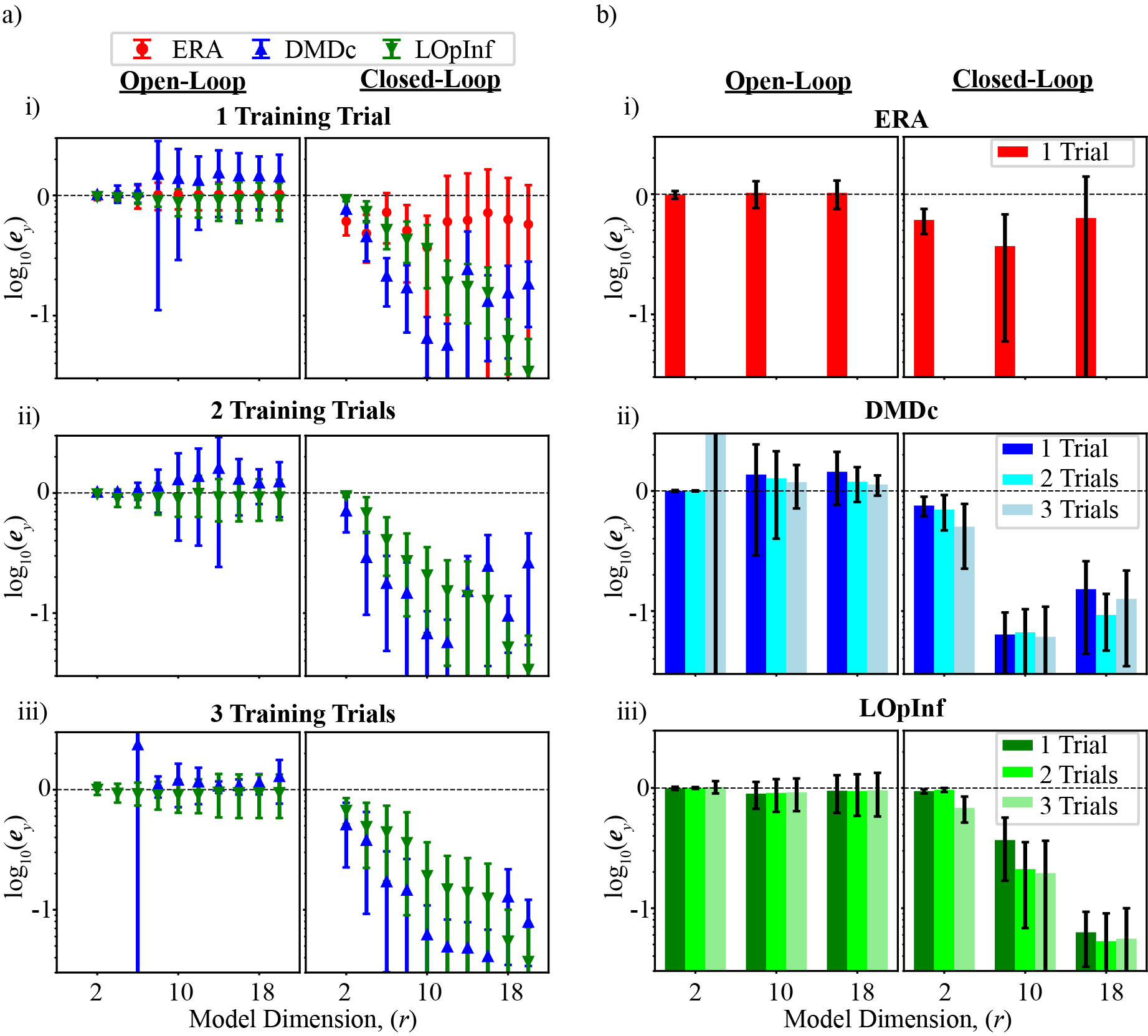}}
\caption{Prediction error of open-loop and closed-loop estimation laws based on ROMs generated via the ERA (red), DMDc (blue), and LOpInf (green) methods. a, i-iii) Comparison across ROM estimation methods of relative estimation error, $e_y$, as a function of ROM dimension, $r$, grouped by the amount of training data used to train ROMs of each method: i) one trial, ii) two trials, iii) three trials. For nearly all synthesized ROMs, except higher dimensional ROMs generated via LOpInf, open-loop relative estimation error over a trial remained approximately at or above $e_y=1$ (black dotted line). Values for DMDc-based ROMs trained on three trials of data and with state dimensions of $r=2$ (${e_y=2946.6\pm3601.7}$) and $r=4$ (${e_y=2150.3\pm2665.9}$) were off the scale and are omitted for visual clarity. In contrast to the open-loop estimation schemes, the closed-loop state observers significantly improve estimation accuracy for nearly all models. b, i-iii) Estimation accuracy of each ROM as a function of the state dimensions tested in our proposed control loops, grouped by ROM synthesis method: i) ERA-based ROMs, ii) DMDc-based ROMs, iii) LOpInf-based ROMs.}
\label{Fig:EstimationAccuracy}
\end{figure*}

We evaluated the open-loop and closed-loop estimation accuracy of each ROM over the 37 test trials specified in Table~\ref{tab:DatasetParameters}, the results of which are summarized in Figure~\ref{Fig:EstimationAccuracy}. For each trial, we computed the relative estimation error $e_{y}$ of each ROM as
\begin{equation}
    e_{y} = \frac{ \lVert\bY-\tilde{\bY}\rVert_F}{\lVert \bY\rVert_F}
\end{equation}
where 
$\bY~=~[\by_0, \cdots, \by_{K-1}]$ is a matrix of ground truth outputs from the simulated soft robot and $\tilde{\bY}~=~[\tilde{\by}_0, \cdots, \tilde{\by}_{K-1}]$ is a matrix of output estimates produced by a given ROM and estimation law. Notably, a value of $e_{y}=1$ coincides with the quality of prediction that would result from an estimation law that only predicts $\tilde{\by}_k=\bzero$ (i.e., the neutral position of the robot).  

We computed $e_{y}$ over $K=1000$ timesteps using both open-loop and closed-loop estimation laws with ROMs generated for each $r=2,4, ..., 20$ and one, two, or three training trials (Fig~\ref{Fig:EstimationAccuracy}a). In open-loop rollouts with each trial, we found that all models produced an average of approximately $e_{y}=1$ for $r=2$. On average, ROMs synthesized using the ERA and LOpInf methods improved in open-loop estimation accuracy as the ROM dimension increased, up to a limit within the tested values of $r$. Additionally, we found that as the average relative estimation error improved, the variance over test trials tended to increase. In contrast, the open-loop estimation error of ROMs generated by DMDc tended to increase with $r$.

In the closed-loop estimation setting, the estimation errors produced by each ROM tended to improve. We found that the values of $e_{y}$ produced by ERA-based and DMDc-based ROMs were lower than those produced by LOpInf-based ROMs for smaller $r$. However, the values of $e_{y}$ generated from ERA-based and DMDc-based ROMs exhibited local optima between reduced-order state dimensions of ${r=10}$ to ${r=16}$, depending on the number of trials used for training. In contrast, the values of $e_{y}$ resulting from LOpInf-based ROMs demonstrated an approximately monotonic improvement with increasing $r$. Additionally, while increasing the amount of training data tended to improve closed-loop estimation accuracy for LOpInf-based ROMs (Fig~\ref{Fig:EstimationAccuracy}b.iii), this tendency was less-so apparent in DMDc-based ROMs (Fig~\ref{Fig:EstimationAccuracy}b.ii).

\section{ROMPC Implementation and Evaluation on Feasible Reference Trajectories}
\label{sec:rompcTests}
We next combined the closed-loop state observer \eqref{eq:estLaw_CL} and model predictive control \eqref{eq:rompc} to evaluate each model reduction method in our proposed ROMPC scheme (Fig.~\ref{blockDiagram}) on our simulated testbench. To implement the control scheme, we developed custom modules in SOFA using a commercially available solver (GUROBI, GUROBI Optimization LLC) for the large-scale quadratic programs that result from applying our ROMPC formulation over long time horizons. Using our simulated testbed, we investigated how ROM dimension and quantity of training data and objective function tuning affects shape tracking performance of the resulting controller. 

We describe this empirical study in this section, beginning with how we generated feasible reference trajectories in Section~\ref{subsec:FeasibleReferenceTrajectoryGeneration}, and then discussing preliminary objective function tuning in Section~\ref{subsec:ObjectiveFunctionTuning}. Finally, we investigate the effects ROM hyperparameters and objective function tuning in Section~\ref{subsec:EffectsOfRomHyperparams} and compare the best-case ROMPC policies generated with each model reduction method in Section~\ref{subsec:bestCaseControllers}.

\subsection{Feasible Reference Trajectory Generation}
\label{subsec:FeasibleReferenceTrajectoryGeneration}
As with many soft robots, our simulated system exhibits highly underactuated dynamics\cite{della_santina_model-based_2023}, necessitating consideration for the dynamic feasibility of any given reference trajectory: infeasible reference trajectories can preclude accurate tracking, making it difficult to assess whether or not poor tracking is due to controller design and tuning or due to the infeasibility of the reference trajectory. In this first set of experiments, we addressed this problem of generating dynamically feasible reference trajectories by using the output trajectories from the eight trials of our dataset denoted by asterisks in Table~\ref{tab:DatasetParameters}---as these output trajectories have already been realized, their feasibility is guaranteed.  

\subsection{Objective Function Tuning}
\label{subsec:ObjectiveFunctionTuning}

Our ROMPC formulation in Section~\ref{sec:rompcFormulation} poses a multiobjective quadratic program where the coefficients of the diagonal matrices, $\bW_\by$, $\bW_\bu$, and $\bW_{\bdu}$, determine the relative importance of reference tracking, magnitude of the control input, and discrete rate of change in the control input, respectively~\cite{borrelli_predictive_2017}. 

We iteratively tuned this objective following multiple heuristics. Because we used reference trajectories that were guaranteed to be feasible and all actuators were identical, we limited selection of $\bW_{\bu}=c_{\bu}\Idm$ and $\bW_{\bdu}=c_{\bdu}\Idm$, i.e., scalar multiples of the identity matrix. Additionally, we selected the diagonal entries of $\bW_{\by}$ that weigh the $z$-displacement tracking error of each control node be a constant $c_z$, with the remaining entries of $\bW_{\by}$ being set to $0$. We applied no tracking penalties to elements of the output corresponding to the $x$-displacements of each control node as we found that doing so could lead to undesirable behaviors (e.g., overly-aggressive control inputs). These undesirable behaviors could potentially be due to the fact that defining both the $x$- and $z$-displacements is redundant given the limited configuration space of the robot and the fixed arc lengths between control points. 

Using the output trajectory realized in Trial~40 in Table~\ref{tab:DatasetParameters} as reference, we iteratively tuned the objective function weights with the heuristic of finding a single set of values for $\bW_\by$, $\bW_\bu$, and $\bW_{\delta\bu}$ that could provide the best control across controllers operating on ROMs from each synthesis method. During tuning, all ROMs were of dimension $r=18$ and trained on three simulation trials for DMDc-based and LOpInf-based ROMs and on one trial of data for the ERA-based ROM (training trial selection is described in Section~\ref{sec:romTraining}).

Our iterative tuning process involved varying the scalar value defining each of the weighting matrices in our objective function with the goal of improving reference tracking as much as possible for a given class of ROM. We then used those same objective function coefficients in trials with another class of ROM, repeating this iteration until tracking performance stopped improving for any controller. Notably, we found that manually tuning the ERA-based controller was the most challenging as small changes to weighting coefficients could result in aggressive or unstable behavior. This tuning process resulted in values of $c_{\bu}=1500$, $c_{\bdu}=1600$, and $c_{z}=0.6$, used for all three ROM classes and dimensions $r$.

\subsection{Effects of ROM Hyperparameters on Tracking Error}
\label{subsec:EffectsOfRomHyperparams}
Using the tuned objective function, we conducted control experiments to determine the tracking performance of our ROMPC scheme as a function of the reduced-order state dimension and amount of training data used to generate the ROMs used for control. To assess tracking performance of various controllers, we selected eight reference trajectories from the 40-trial dataset presented earlier (denoted by asterisks in Table~\ref{tab:DatasetParameters}). These trajectories were selected to represent a variety of dynamically feasible behaviors. 
\subsubsection{Error Metrics}
\label{subsubsec:ErrorMetrics}
To quantify the tracking performance, we considered two error metrics. First, we considered the relative tracking error of the robot's full body over the entirety of a control trial, 
\begin{equation}
    e_{r} = \frac{ \lVert\bY^*-\bY\rVert_F}{\lVert\bY^*\rVert_F} ,
    \label{eq:relTrackingError}
\end{equation}
where $\bY^*=[\by^*_{0}, \cdots, \by^*_{K-1}] \in \mathbb{R}^{p\times K}$ is reference output trajectory for a given trial and ${\bY=[\by_{0}, \cdots, \by_{K-1}]\in \mathbb{R}^{p\times K}}$ is the realized output trajectory for the same trial. Notably, a value of $e_{r}=1$ indicates a quality of tracking that would result from a ``do-nothing'' policy (i.e., one that produces ${\bu_k=\bzero}$ for all $k$).
We also considered a pointwise relative tracking error
\begin{equation}
    e_{r,j} = \frac{ \lVert\bY^*_j-\bY_j\rVert_F}{\lVert\bY^*_j\rVert_F}
    \label{eq:spatialError}
\end{equation}
where ${\bY^*_j=[y^*_{j,0}, \cdots, y^*_{j,K-1}] \in \mathbb{R}^{K}}$ are the reference (target) outputs at control point $j$ and ${\bY_j=[y_{j,0}, \cdots, y_{j,K-1}]\in \mathbb{R}^{K}}$. Similar to the definition of $e_r$, a value of $e_{r,j}=1$ indicates a tracking quality at point $j$ equivalent to that which would result from a control policy that keeps that point stationary, such as the same ``do-nothing'' policy described previously.
\begin{figure*}[htbp]
\centerline{\includegraphics[width=6.5in]{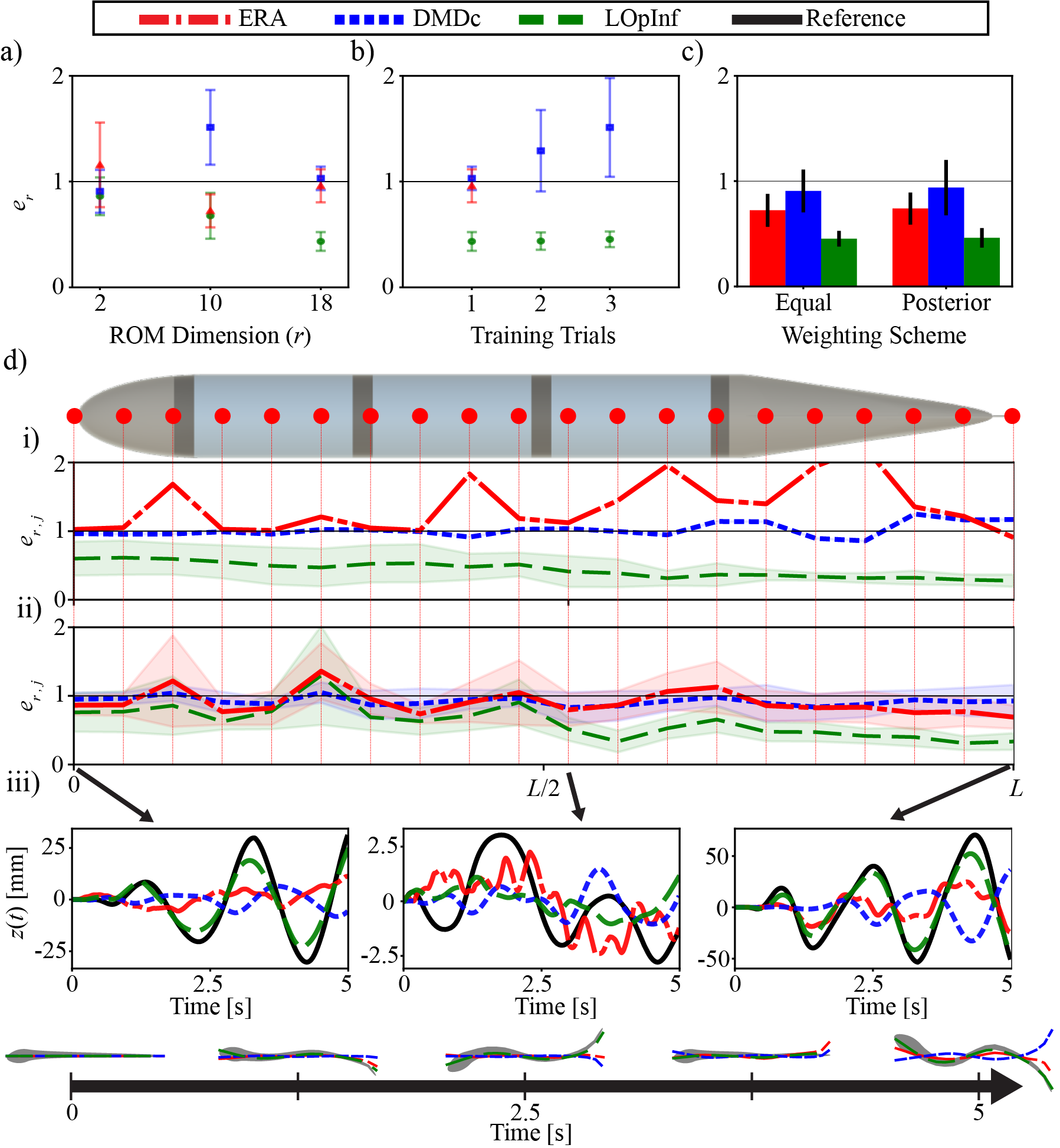}}
\caption{Tracking performance of ROMPC scheme over ROM types, dimensions, quantities of training data, and objective function tuning schemes. The average tracking error for ERA-based (red), DMDc-based (blue), and LOpInf-based (green) controllers generated through this process followed from an extensive tuning process that considered the effects of (a) reduced-order dimension, $r$, for ROMs trained on one trial of data and (b) quantity of training data used for generated ROMs of dimension 18 on tracking performance. (c) Our tuning process also explored the effects of varying coefficients in the MPC objective function to apply equal weighting to tracking penalties across the body as well as weights defined by a posterior-focused scheme. (d) Relative tracking error of the best observed controllers based on each ROM generation method computed over the robot's body length. Tracking error was computed over known feasible trajectories from (i) the trials used to train each ROM and (ii) a set of eight test trials from our dataset. The average tracking error for each best-case controller over each set of trials is shown as dotted lines, with surrounding shaded regions representing one standard deviation of tracking error at a given point. (iii) Output trajectories generated by each controller are shown at control points located at the head (top left), midpoint (top middle), and tail (top right) of the robot for an example reference trajectory (black) extracted from Trial~33 of our dataset. The full reference centerline (black) of the simulated robot are shown for the same trial (bottom) overlaid with centerlines produced by each best-case controller shown in their respective colors (deflections in the $z$-direction amplified by a factor of five for sake of visualization).}
\label{fig:controlResults_NumericalFeasTraj_EqualWeighting}
\end{figure*}
\subsubsection{Tracking Performance as a Function of Reduced-Order State Dimension}
We first evaluated the tracking performance of our ROMPC scheme as we varied the value of $r$ used to generate each ROM (Fig.~\ref{fig:controlResults_NumericalFeasTraj_EqualWeighting}.a). For this experiment, we considered ROMs of dimension 2, 10, and 18 and used models trained only on Trial 40 in Table~\ref{tab:DatasetParameters} when computing $e_r$ over the eight selected reference trajectories. 

Within the tested values of $r$, the values of $e_r$ produced by the LOpInf-based controllers were either on par with, or significantly lower than those produced by DMDc-based and ERA-based controller. Moreover, we observed a monotonic decrease, on average, in $e_r$ with increasing $r$ for LOpInf-based controllers. In contrast, DMDc-based and ERA-based controllers exhibited local minima in tracking error at values of $r=10$ and $r=2$, respectively. Regardless, the best-case DMDc-based and ERA-based controllers in this experiment still exhibited significantly higher full-body tracking error than the best-case LOpInf-based controller.

\subsubsection{Tracking Performance as a Function of ROM Training Data Quantity}
Next, we assessed how tracking performance varied as a result of the amount of data used to train each ROM (Fig.~\ref{fig:controlResults_NumericalFeasTraj_EqualWeighting}.b). For this experiment, we evaluated the tracking error $e_r$, produced by ROMs generated from the same sets of one, two, and three training trials presented previously, all of the same dimension, $r=18$. Due to the formulation of ERA-based ROMs used here, we only assessed the performance of the ERA-based controller when the ERA-based ROM was trained on one trial of data. 
    
We found that the LOpInf-based controllers produced approximately constant values of $e_r\approx0.45\pm0.08$ over all ROMs tested. The DMDc-based and ERA-based controllers produced tracking errors of $e_r\approx1$ for models trained on one trial of data. Notably however, the tracking performance of DMDc-based controllers degraded as the amount of data used for training increased. 

\subsubsection{Tracking Performance as a Function of Varying Tracking Penalty Weights}

Finally, we considered how tracking performance was affected by varying the scheme used to define weighting coefficients in $\bW_\by$ for tracking the reference $z$-displacements at control nodes across the soft robot's body (Fig.~\ref{fig:controlResults_NumericalFeasTraj_EqualWeighting}.c).
For this experiment, we considered two weighting schemes: 1) an equal weighting of tracking penalties across the robot's body (i.e., the same scheme as described previously); and 2) a posterior-focused weighting scheme where tracking penalties were only non-zero for the posterior half of the robot. For the latter weighting scheme, $\bW_\by$ was still kept as a diagonal matrix wherein the only nonzero diagonal values were those applying penalty to the $z$-displacement tracking error for the 10 posterior-most control points along the simulated robot's centerline. Using values of $1.2$ for these nonzero entries of $\bW_\by$, we found that, relative to the equal weighting scheme, the posterior-focused scheme slightly increased the average and standard deviations of the full-body tracking errors resulting from the DMDc-based and LOpInf-based controllers, while negligibly affecting the average tracking performance of the ERA-based controller.

\subsection{Comparison of Best-Case Controllers}
\label{subsec:bestCaseControllers}

Based on the results from the previous section, we compared the pointwise tracking errors resulting from the best-case controllers representing each ROM synthesis technique in our comparative study (Fig.~\ref{fig:controlResults_NumericalFeasTraj_EqualWeighting}.d). The ERA, DMDc, and LOpInf ROMs used were of dimension 10, 2, and 18, respectively, and the ERA and DMDc ROMs were each trained on one trial of data while the LOpInf ROM was trained on 3 trials of data. 

We first considered how each controller tracked the reference trajectories from trials on which their respective models were trained (Fig.~\ref{fig:controlResults_NumericalFeasTraj_EqualWeighting}.d.i). On these training trials, we found that over the entirety of the robot's body, the tracking errors of the DMDc-based and ERA-based controllers were approximately equal to, or higher than $e_{r,j}=1$, qualitatively indicating performance that was on par with, or worse than, a ``do-nothing'' policy. In contrast, the best-case LOpInf-based controller consistently performed better than a ``do-nothing policy" with an average tracking error of $e_{r,0}=0.60$ at the head of the robot and a value of $e_{r,19}=0.28$ near at the tail. 

We then considered the pointwise relative tracking errors on the eight test trials used in our prior estimation studies (Fig.~\ref{fig:controlResults_NumericalFeasTraj_EqualWeighting}.d.ii). Notably, the best-case ERA-based and DMDc-based controllers generally exhibited \textit{better} pointwise tracking performance on these test trials when compared to their tracking performance with reference trajectories from the training trials discussed previously. In contrast, the tracking performance of the best-case LOpInf controller degraded slightly near the anterior end of the robot. Furthermore, when considering the timeseries $z$-displacement for each of the control points, we found that at the anterior and posterior ends of the robot, the best-case LOpInf-based controller typically exhibited tracking performance that best matched the behavior of reference trajectories (Fig.~\ref{fig:controlResults_NumericalFeasTraj_EqualWeighting}.d.iii). In contrast, the best-case ERA-based and DMDc-based controllers often generated output trajectories with irregular oscillations at much higher frequencies than those of the oscillations prescribed by the reference trajectories. 

\section{Control Experiments with Increasingly Complex Reference Trajectories}
\label{Sec:ControlExperiments}
Using our simulated testbed and the best-case controllers described in Section~\ref{subsec:bestCaseControllers}, we conducted two control experiments wherein the simulated robot tracked full-body reference trajectories generated from systems of increasing complexity and realism. The first experiment, described in Section~\ref{sec:controlExperiment1}, focused on tracking trajectories generated from an empirical kinematic model of anguilliform swimming. The second experiment, described in Section~\ref{sec:controlExperiment2}, used reference trajectories extracted from experiments with a reduced-scale physical analog of our simulated robot. For each experiment, we describe how we generated reference trajectories as well as the tracking performance of the best-case controllers representing each model reduction method.

\subsection{Control Experiment 1: Tracking Bioinspired Full-Body Trajectories}
\label{sec:controlExperiment1}

We first evaluated how well the simulated platform can track the traveling wave kinematics integral to natural anguilliform locomotion during straight-line swimming~\cite{tytell_hydrodynamicsII_2004}. 
Along with being a key element of anguilliform swimming, traveling wave propagation through a finitely actuated elastic structure is a uniquely dynamic phenomenon~\cite{tytell_hydrodynamicsII_2004}.
Thus, the goal of this control experiment to characterize how well the simulated system could track the gaits generated by this dynamic process. 

\subsubsection{Reference Trajectory Generation}
We used the empirically derived model of centerline kinematics presented in \cite{tytell_hydrodynamicsII_2004}, that describes the lateral position $z_m$, of an eel's centerline at time $t$, as
\begin{equation}
    z_m(s,t) = A~e^{\alpha(\frac{s}{L}-1)} \sin(ks - 2\pi f t),
\end{equation}
where $s\in[0,L]$ indicates the axial position along the robot's centerline (with a value of $s=0$ corresponding to the anterior-most point), $L$ is the length of robot, $A$ is the tail beat amplitude, $k$ is the body wavenumber, $f$ is the tail beat frequency, and $\alpha$ is the amplitude growth rate \cite{tytell_hydrodynamicsII_2004}. 

We selected a broad set of gait parameters to generate reference trajectories from Table~\ref{tab:BioinspiredGaitParams}, including values that have been empirically observed in anguilliform animals~\cite{tytell_hydrodynamicsII_2004}. For each of the 12 combinations of gait parameters shown in Table~\ref{tab:BioinspiredGaitParams}, we simulated tracking with each best-case ROMPC controller found in the previous section over a duration of 10~seconds (i.e., 1000~time steps). We quantified tracking performance with the same error metric defined in \eqref{eq:relTrackingError}, considering only the tracking performance after the first 5~seconds of each trial. We ignored this transient portion of each trial to only account for the steady-state tracking performance of each controller. 

\definecolor{Nobel}{rgb}{0.7529,0.7529,0.7529}
\begin{table}
\caption{Parameters defining tracking experiments with bioinspired gaits.}
\label{tab:BioinspiredGaitParams}
\centering
\begin{tblr}{
  width = \linewidth,
  colspec = {Q[410]Q[475]},
  cells = {c},
  row{1} = {Nobel},
  hlines,
  vlines,
}
Gait Parameter & Parameter Values  \\
Tail beat amplitude, $A$ [mm]         & $30$        \\
Growth rate, $\alpha$          & $1.0, 3.5$             \\
Tail beat frequency, $f$ [Hz]         & $0.5, 1.0$      \\
Body wavenumber, $k$              &$ 0.5, 1.0, 1.5$ 
\end{tblr}
\end{table}

\subsubsection{Tracking Performance}

\begin{figure*}[htbp]
\centerline{\includegraphics[width=6.5in]{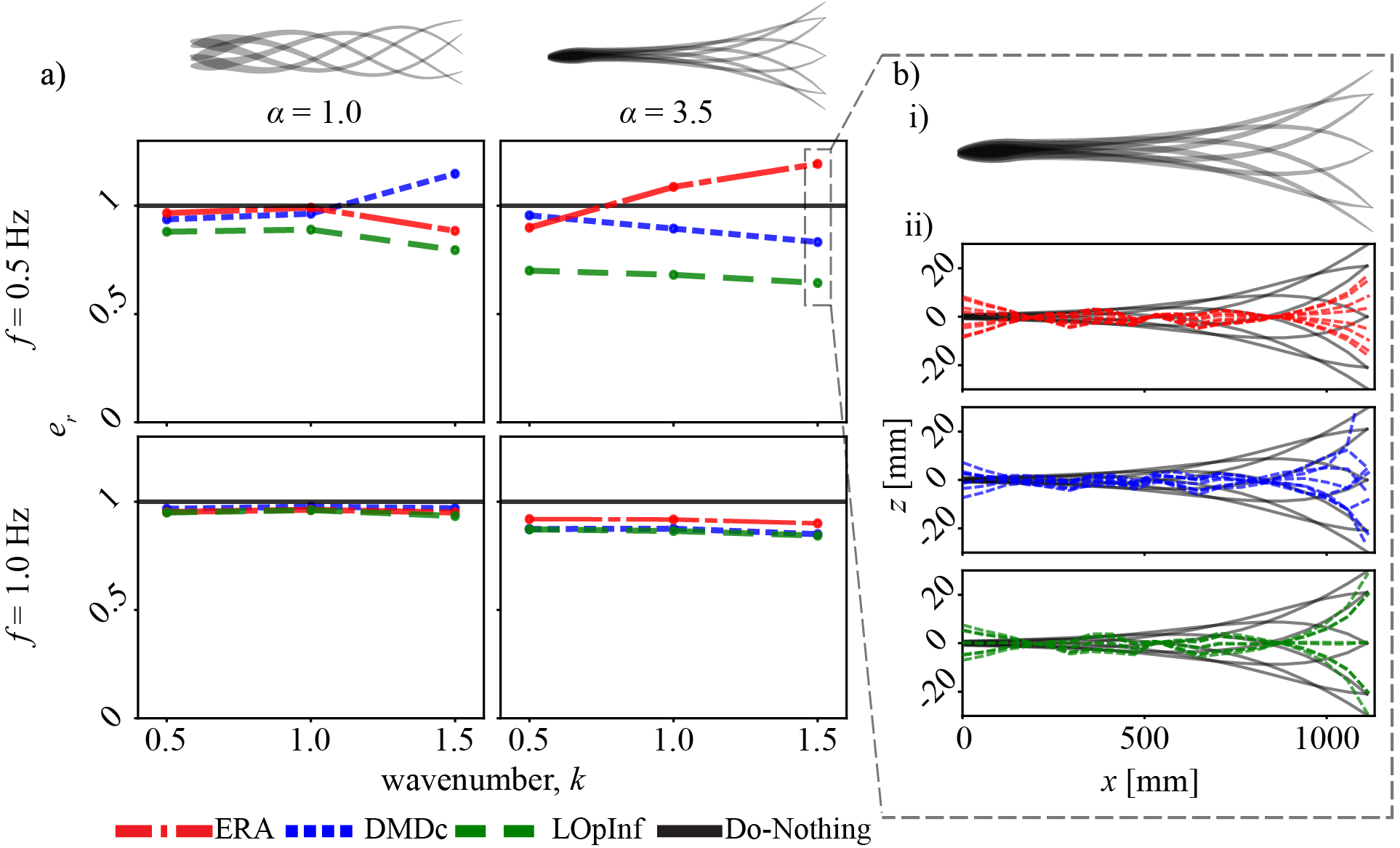}}
\caption{Bioinspired trajectory tracking performance. a) Full-body relative tracking error of robot over all combinations of bioinspired gait parameters. Relative tracking errors were computed as defined in \eqref{eq:relTrackingError}. Results are separated by gait frequencies $f$ (rows) and growth rates $\alpha$ (columns) with example centerlines produced with a given $\alpha$ shown above each column. b) Centerlines of robot produced over a period of steady-state operation using gate parameters $f=0.5$~Hz, $k=1.5$, $\alpha=3.5$, and $A=30$~mm. Centerline snapshots from the reference trajectory (i) for this gait are shown as black lines on each subplot and dotted color lines in (ii) show the centerline snapshots for the best-case ERA-based (top), DMDc-based (middle), and LOpInf-based (bottom) controllers.}
\label{bioinspiredTrajectoryTrackingResults}
\end{figure*}

For all gait parameters tested, the LOpInf-based controller produced the lowest tracking error, though the difference in tracking errors was nearly negligible in trials where $\alpha=1.0$ and $f=1.0$~Hz (Fig.~\ref{bioinspiredTrajectoryTrackingResults}.a). For most sets of parameters, all controllers demonstrated tracking performance better than that of a ``do-nothing'' control policy. We also found that under the DMDc-based and LOpInf-based ROMPC policies, tracking performance often significantly improved for the higher growth rate tested. This could potentially be attributed to the fact that larger values of $\alpha$ significantly reduces the desired amount of lateral movement near the anterior end of the simulated robot (Fig.~\ref{bioinspiredTrajectoryTrackingResults}.a), and this reduced lateral movement better aligns with the boundary conditions we applied in our structural simulation that force lower displacement near the base of the robot's head. 

To further analyze the tracking behavior of our system, we considered the simulated robot's kinematics over a single cycle of operation (Fig~\ref{bioinspiredTrajectoryTrackingResults}.b). We found that, over the selected gait parameters, the midbody of the robot often exhibited the least movement while the anterior and posterior ends of the robot contributed most significantly to the tracking performance. Also, all controllers typically exhibited the lowest tracking error near the posterior end of the robot, with the LOpInf-based controller consistently showing superior tracking performance at this point. The consistent improvement in tracking at the tail of robot can likely be attributed to the fact that the generated reference gaits prescribe the largest lateral displacement at the tail of the robot. This means that during control optimization, reducing error at the tail has the largest potential for reducing the loss prescribed by our objective function.

\subsection{Control Experiment 2: Tracking Trajectories Generated by Experimental Robot}
\label{sec:controlExperiment2}
In our second set of control experiments, we assessed the capacity of the simulated robot to reproduce dynamic swimming behaviors exhibited by the experimental platform developed in \cite{cervera-torralba_lost-core_2024, park_analysis_2025}. The physical system from which data was collected was comprised of the same materials described in Section~\ref{sec:SoftRobotSim}, but is one-third the scale of our simulated system and has minor modifications to enable tethered hydraulic actuation. Further details of these modifications and the system's physical realization can be found in \cite{cervera-torralba_lost-core_2024} and \cite{park_analysis_2025}.  

\subsubsection{Reference Trajectory Generation}
\begin{figure*}[htbp]
\centerline{\includegraphics[width=6.5in]{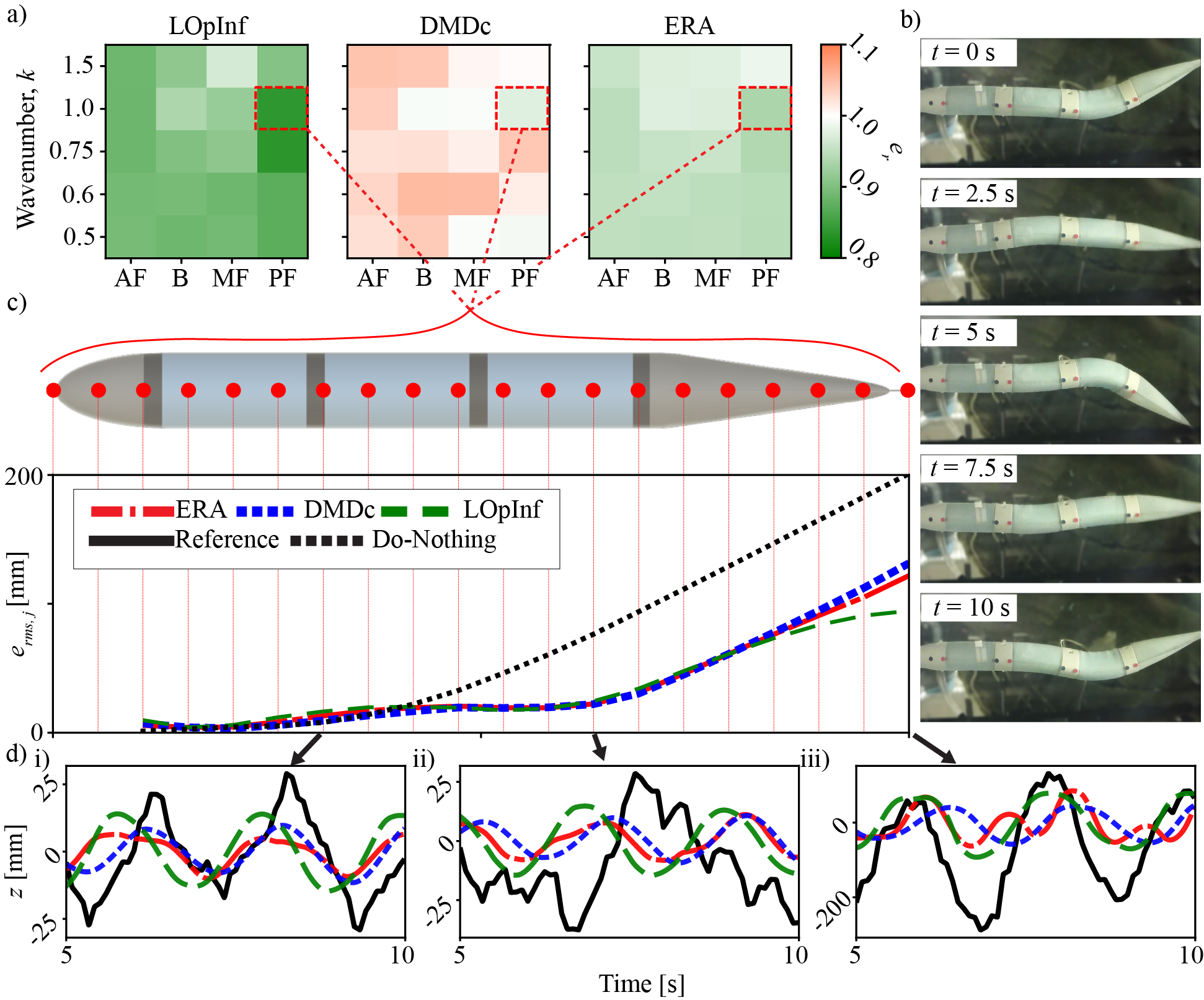}}
\caption{Tracking reference trajectories extracted from experiments with physical robot platform. Using the data collected from the 20 physical experiments conducted in \cite{park_analysis_2025}, we constructed reference trajectories with which we tested the tracking performance of our simulated system with the best-case controllers found with each ROM synthesis method. (a) Full-body tracking error maps resulting from tracking experiments over five gait wavenumbers and four sets of input amplitudes that are either anterior focused (AF), balanced (B), middle focused (MF), and posterior focused (PF). Design and selection of gait parameters are detailed in \cite{park_analysis_2025}. (b) Images of physical robot during a trial with wavenumber $k=1.0$ and posterior-focused input amplitudes. (c) We computed the pointwise relative error over the last 5~seconds of tracking this trial for each controller and (d) show the output trajectories at three points along the robot's body.}
\label{fig:ControlStudy2}
\end{figure*}

To generate reference trajectories for the simulated robot, we extracted the centerline trajectories produced by the physical robot over the 20 thrust-characterization experiments conducted in \cite{park_analysis_2025}. These 20 experiments implemented open-loop control policies designed to approximate anguilliform swimming with a tail beat frequency of 0.5~Hz, nominal wavenumbers of $k \in \{0.5,~0.6,~0.75,~ 1.0,~1.5\}$, and inputs magnitudes meant to generate anterior-focused (AF), balanced (B), middle-focused (MF), and posterior-focused (PF) bending~\cite{park_analysis_2025}. The open-loop control policies used in~\cite{park_analysis_2025} were designed assuming deterministic, quasi-static, and constant-curvature deformation of each actuated segment, resulting in deformations like that shown in Figure~\ref{fig:ControlStudy2}.b. During these physical experiments, the robot's head segment was held fixed, which enforced similar boundary conditions to the simulated platform where the head is allowed to pivot about its base as described in Section~\ref{Subsec:SOFASim}. 

For each of the physical experiments in \cite{park_analysis_2025}, the robot's centerline kinematics were recorded via visual tracking of colored markers placed on each of the rigid couples of the robot. For each video frame, we used these colored markers to estimate the bend angle of each segment, which we then used to fit constant-curvature arcs across the centerline of the robot. We observed that the oscillation frequency of the system was too slow to induce significant bending in the passive tail. So we approximated the configuration of the tail as a line segment extending from the posterior-most actuated segment of the robot. 

While our simulated control loop operated at a frequency of 100~Hz, the sample rate of the visually extracted centerlines was 12~Hz. To address this discrepancy, we linearly interpolated the extracted centerlines over time to produce reference trajectories suitable for our control loop. Finally, we spatially discretized the estimated centerlines into constant-length curves, with the end-points of these curves corresponding to the desired location of the simulated platform's control points. As the experimental data was only collected from the base of the physical robot's head to its tail, we discretized these curves into 17 segments to generate reference trajectories for control points 2-19 (as labeled in Fig~\ref{fig2}).  

\subsubsection{Tracking Performance}
Since the two anterior-most control points were fixed on the physical platform, we modified the cost function for the control optimization to enforce 0 tracking penalty on the $z$-displacement error at these points, keeping the remaining coefficients of the cost function as defined in Section~\ref{subsec:bestCaseControllers}. As before, we considered the full-body relative tracking error metric defined in (\ref{eq:relTrackingError}), ignoring the contributions of the two fixed control points, and again only considering the last 5~seconds of each simulated experiment to reduce the effect of transient behaviors expected from the simulated platform starting from rest. We also considered a pointwise root mean square (RMS) error across the length of the simulated robots body, defined as 
\begin{equation}
    e_{\text{RMS},j} = \frac{ \lVert\bY^*_j-\bY_j\rVert_F}{\sqrt{K-1}},
    \label{eq:spatialErrorRMS}
\end{equation}
We considered this RMS error in lieu of the pointwise relative error $e_{r,j}$, introduced in \eqref{eq:spatialError}, because these experiments prescribed zero (or approximately zero) displacement anterior control points, meaning that even slight lateral movement from the anterior control points of our simulated platform could produce extreme values of $e_{r,j}$, making it difficult to compare tracking quality between controllers.

In considering $e_{\text{RMS},j}$, we found that the RMS tracking error near the posterior end of the robot often demonstrated the largest difference in tracking performance between each controller (Fig.~\ref{fig:ControlStudy2}.c). Even though the DMDc-based and ERA-based controllers sometimes exhibited tracking performance better than that of the best-case LOpInf-based controller near the anterior end of the robot, this difference was always minor in comparison to the improved tracking performance of the LOpInf based controller at the robot's posterior end. 

\begin{figure}[tb]
\centerline{\includegraphics[width=3.0in]{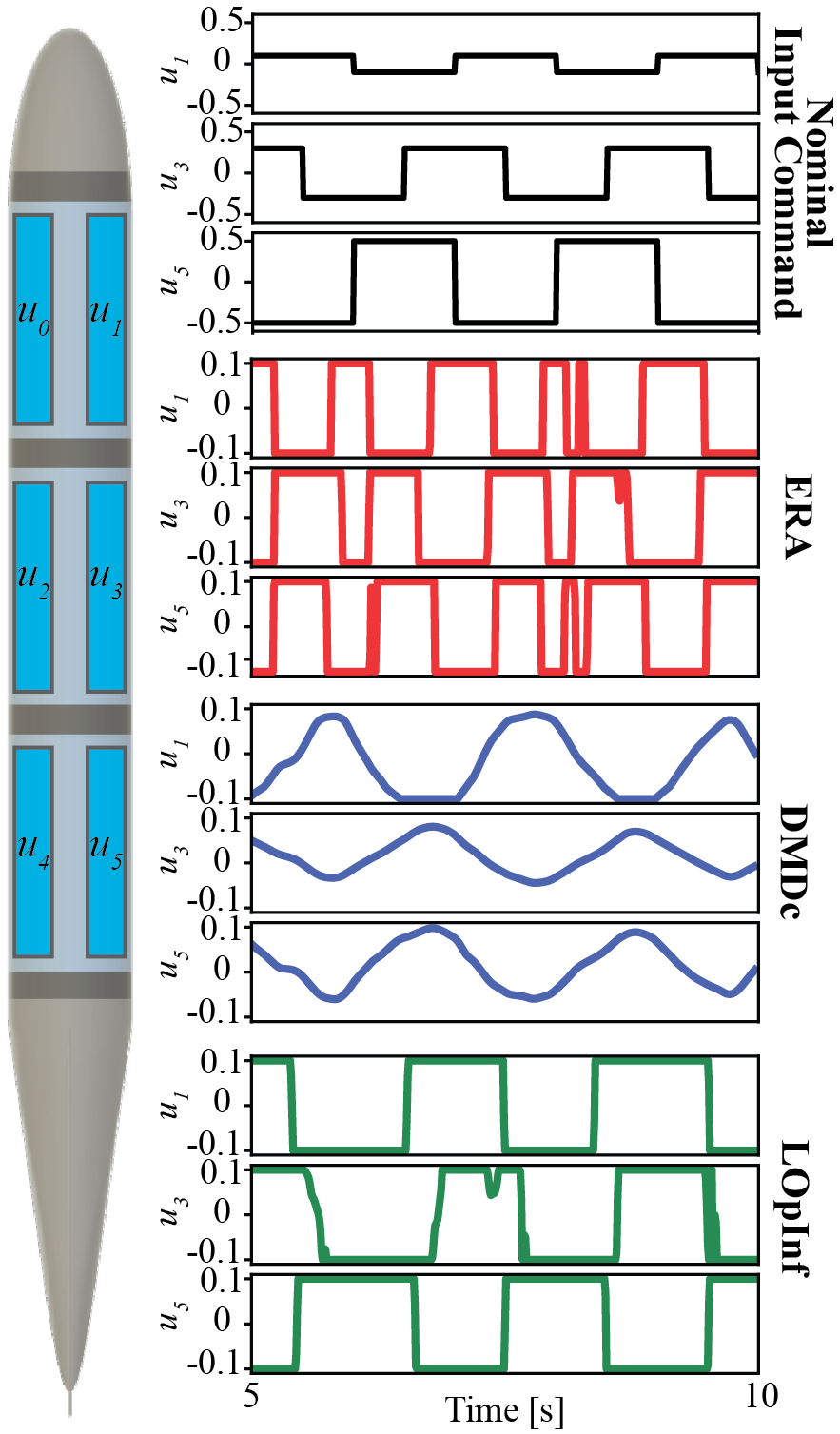}}
\caption{Control signals produced by simulated controller when tracking reference trajectories generated from experiment with the physical robot in~\cite{park_analysis_2025} with wavenumber, ${k=1.0}$ and (PF) bending amplitudes. The nominal input commands (black) used for the physical experiment represent the control signals sent from the microcontroller controlling the physical system to the electronic speed controllers driving the pumps connected to each actuator. We normalized these commands such that the maximum peak-to-peak amplitude of the command signal that the microcontroller produced has a value of $1.0$. As these are the open-loop signals generated by the microcontroller, they do not account for the dynamics and delays induced by the electrical drivers, pumps, and fluidic channels of the physical system and are not pressures. The non-dimensionalized pressure control signals produced by the ERA-based (red), DMDc-based (blue) and LOpInf-based (green) policies for inputs $u_1$, $u_3$, and $u_5$ were fed directly into the full-order simulation.}
\label{fig:physExp_inputs}
\end{figure}

For all trials, all controllers followed a similar trend in RMS tracking error over the length of the robot's body---RMS tracking error remained relatively low up to the posterior half of the robot, after which the error tended to increase more rapidly. We believe this behavior is best explained through the timeseries responses of each control point under each controller (Fig.~\ref{fig:ControlStudy2}.d). Tracking performance most heavily degraded when the desired magnitude of the reference became large---a phenomenon that always happened near the posterior of the robot. We contend that large deformations led to degraded tracking because the maximum pressure constraints applied to our simulated system were likely lower than what was necessary to produce the bending prescribed by the reference trajectories. Simulating pressures beyond these saturation constraints negatively impacted simulation stability, limiting our capacity to reduce tracking error by expanding the saturation window. However, we validated this hypothesis by observing that the input trajectories produced by each controller would frequently saturate (Fig.~\ref{fig:physExp_inputs}). Further inspection of the input trajectories to each actuator showed that ERA-based controller often switched seemingly sporadically while the DMDc-based and LOpInf-based controllers oscillated at the expected input frequency of 0.5~Hz, with the LOpInf controller producing control inputs closest to the square-wave nominal signal used by the physical system, though with phase shifts. 

\section{Discussion}
\label{sec:Discussion}
The primary objective of this comparative study is to quantitatively and qualitatively compare representative methods for data-driven model reduction in an identical setting requiring dynamic shape control of a soft robot. In this section, we use the results presented thus far to compare each model reduction method in Section~\ref{subsec:ComparisonofRomMethodsForROMPC}. We then describe various limitations of this study in Section~\ref{subsec:limitations}.
\subsection{Comparison of Model Reduction Methods for ROMPC-based Shape Control}
\label{subsec:ComparisonofRomMethodsForROMPC}
Two common heuristics guiding the use of data-driven models for control are 1) an expected tradeoff between model fidelity and computational cost, often linking higher-dimensional models with more accurate simulations; and 2) an expectation that increasing the amount or quality of training data will increase or maintain model accuracy~\cite{armanini_soft_2023, haggerty_control_2023}. Within the evaluated range of ROM state dimensions $r$, our estimation experiments in Section~\ref{sec:romTraining} demonstrated that the LOpInf-based models best matched these expectations in both open-loop and closed-loop prediction settings. This behavior could be attributed to the second-order mechanical structure enforced by LOpInf models. Provided sufficiently expressive data, the Lagrangian form described by equation \eqref{eq:lopinf} has been shown to effectively capture the dominant dynamics of a soft system with purely linear dynamics~\cite{sharma2024preserving}, producing low open-loop prediction errors over long time horizons. As our simulated system integrates geometric nonlinearity, we expected that the linear ROMs produced via LOpInf would not capture these nonlinearities. However, compared to the DMDc and ERA based models, the LOpInf-based models nonetheless produced the most accurate open-loop estimates on average for every $r$ except for when $r=2$ and models were trained on only one trial of training data. 

In the closed-loop estimation setting, the ERA-based and DMDc-based ROMs often produced lower estimation error than LOpInf-based ROMs for $r\leq12$ when one or two trials of training data were used and $r\leq18$ when three trials of training data were used. However, the DMDc-based and ERA-based models would exhibit sudden increases in estimation errors as the reduced-order state dimension increased beyond these values. We hypothesize that this behavior could be due to the fact that the DMDc algorithm presented in~\cite{proctor_dynamic_2016} extracts dominant spatial modes presented in the data and constructs a system representation where the state corresponds to a linear combination of those spatial modes. Therefore, it might be possible that by learning from datasets that inappropriately over-represent certain spatial modes, the prediction quality of the model can degrade even if a higher state dimension or more data are used during training. In a similar vein, as the ERA-based models could only learn from a single trial of data, albeit generated with rich dynamics, it is possible that the trial we selected only effectively represented a limited set of observable modes amenable to closed-loop estimation. Increasing state dimension to include further modes could therefore have degraded closed-loop estimation accuracy. 

When comparing the ROMs in our proposed ROMPC scheme, we initially hypothesized that ROMs producing the most accurate state reconstruction would admit controllers with the most accurate tracking. This hypothesis was based on the idea that the control optimization in (\ref{eq:rompc}) requires an accurate estimate of the current state, which is provided by the closed-loop state observer in our control loop. While we found that this hypothesis held true in the best-case ERA-based and LOpInf-based controllers, it did not hold true for the best-case DMDc-based controller. In fact, the best DMDc-based controller was based on a ROM trained on one trial of data with reduced-order state dimension $r=2$, which notably produced the worst-case closed-loop estimation error among all hyperparameters tested. 

Along with enabling a comparison between three classes of data-driven model reduction techniques in a control setting, our comparative approach additionally provided insights into the design and operation of the anguilliform robot that we simulated. When considering the pointwise tracking errors that resulted from tracking dynamically feasible reference trajectories, we found that all of the best-case controllers tended to exhibit spikes in tracking error just before each of the rigid couples comprising the simulated robot (Fig.~\ref{fig:controlResults_NumericalFeasTraj_EqualWeighting}.d.ii). This consistent pattern could indicate a potential reduction in either control authority or model fidelity (or both) near these regions, eliciting either design iteration or model refinement (or both) to improve tracking performance.

One of the most notable results from the control experiments in Section~\ref{Sec:ControlExperiments} is that, even when tracking reference trajectories that are not known \textit{a priori} to be dynamically feasible, the best-case LOpInf-based controller consistently produced superior tracking over the robot's body. This performance is best explained by the consistently lower tracking error near the tail of the robot, where reference displacements were often the highest and, therefore, of the highest priority to track given the ROMPC objective function. However, as the tail is unactuated, effectively tracking dynamic reference trajectories in this section of the robot indicates that the LOpInf-based controller is best accounting for the underlying structural dynamics of the robot during each trial. Notably, the best-case LOpInf-based controller produced this superior performance with an objective function that we had originally tuned to be identical for all of the ROMs tested. An objective function tuned specifically for a ROMPC policy based on a specific ROM instance will produce tracking errors that are equal to, or better than the ones shown in this work; we refrained from doing this to better compare the three model classes.

Moreover, the results from tracking bioinspired reference trajectories in Section~\ref{sec:controlExperiment1} illustrated that the selection of gait parameters can significantly affect tracking performance. This is attributed to some selections of gait parameters producing reference trajectories that are closer to being feasible than those reference trajectories produced by other gait parameters. Nonetheless, none of these reference trajectories were expected to be fully feasible due to differences between our simulated platform and the physical systems from which reference trajectories were derived, as explained next. 

\subsection{Limitations}
\label{subsec:limitations}
Even though the LOpInf-based controller produced superior tracking performance in all of the experiments with feasible trajectories (Section~\ref{sec:rompcTests}), it regularly produced the highest pointwise relative tracking error in more anterior portions of the simulated robot's body---a trend that was also often apparent in the DMDc-based and ERA-based controllers, though less pronounced. In the setting of tracking feasible reference trajectories, this tracking error could be in part attributed to controller tuning, but it more likely stemmed from the fact that our ROMPC policy leveraged a single linear ROM to represent the dynamics of an otherwise nonlinear full-order system. Nonetheless, we considered this error a baseline for our ROMPC scheme against which we compared the results of the experiments that utilized infeasible reference trajectories. 

Compared to this baseline, the experiments in Section~\ref{Sec:ControlExperiments} regularly exhibited a higher full-body tracking error for all models. We hypothesize that the higher tracking errors in these experiments can be attributed to one of two sources. First, the reference trajectories used in these studies were generated from systems that involved forcing from FSI that were not included in our simulated testbench or datasets. Even though all of the model reduction methods used here could be made to account for FSI, our full-order model could not represent these effects due to current limitations in modern simulation tooling for fluidically driven soft robots operating underwater. A second potential source of tracking error beyond the aforementioned baseline may have stemmed from structural and other operational differences between the simulated platform and the physical systems from which reference trajectories were generated. Along with any differences in shape between our simulated robot, the eels from which trajectories were modeled in \cite{tytell_hydrodynamicsII_2004}, and the physical platform in \cite{park_analysis_2025}, there were additional discrepancies in boundary conditions between all of these systems. These differences are of particular significance to consider for the data-driven ROMs produced via DMDc and LOpInf, as the reduced-order states generated by these methods represent dominant spatial modes present in the training data. Because all of the training data in this work were produced from simulations with boundary conditions that allow the head to pivot about the constraints described in Section~\ref{Subsec:SOFASim}, the LOpInf and DMDc methods can only project the full-order data onto bases where this pivoting is represented. As a result, the generated ROMPC policies are limited in their capacity to account for alternative boundary conditions, such as those seen in Section~\ref{sec:controlExperiment2} where the entire head of the physical system is held stationary.

\section{Conclusions and Future Work}
\label{sec:Conclusion}
We presented a comparative study of linear, data-driven model reduction techniques in the setting of dynamic shape control for soft robots. We selected three methods---ERA, DMDc, and LOpInf---representing different families of data-driven model reduction techniques and compared them on a custom simulated testbench of the structural dynamics of an anguilliform-inspired soft robot. Using this testbench, we generated a large-scale dataset that we used to assess the open-loop and closed-loop estimation accuracy of ROMs generated by each method as a function of ROM dimension and amount of training data. This dataset is publicly available at~\cite{Adibnazari2025Dataset}. Finally, we compared each class of ROM in multiple control settings---one where full-body reference trajectories were guaranteed to be dynamically feasible, one where reference trajectories came from previously established bioinspired models of anguilliform locomotion, and one where the reference trajectory came from a reduced-scale physical analog of our simulated platform. 

In this work, we in part explored reference trajectories that were guaranteed to be feasible as they had previously been simulated. The broader problem of efficiently generating feasible full-body trajectories for a given soft robot and task remains largely unanswered in the literature. However, this question is of critical importance for continued work on dynamic shape control of soft and hybrid-soft robots. This problem is challenging because generating feasible trajectories for a given task requires an accurate dynamic model of the system, which is fundamentally precluded in our setting by the lack of available tooling for simulating fluid-structure interaction in our full-order model. Therefore, future extensions of this work would benefit most from the development of accurate full-order three-dimensional FEM models that incorporate fluid-structure interaction. Provided such full-order models, future work could then leverage data-driven model reduction techniques such as LOpInf to produce low-dimensional surrogate models with which feasible---or approximately feasible---reference trajectories could be generated online for a given task. 

Additionally, a valuable avenue for further research on ROMPC-based shape control of soft robots could focus on accounting for various nonlinear effects, such as geometric nonlinearity and nonlinear material behaviors, as well as other sources of modeling error in a manner that is still amenable for online control. Accounting for these effects in both the simulation and data collection phase as well as in the modeling capacity of any synthesized ROMs is important for improving the performance of the resulting ROMs in both estimation and control settings. One potential approach relevant to the DMDc-based and LOpInf-based controllers could focus on developing alternative methods for identifying relevant bases onto which full-order state is projected during model synthesis. Another approach could leverage various extensions of the LOpInf method that account for nonlinear dynamics present in the full-order model while preserving the physics-based structure of the system~\cite{sharma2024lagrangian, sharma2025nonlinear}; using these methods in an online dynamic shape control setting would then require the use of nonlinear model-based control techniques that appropriately incorporate the resulting ROMs.      

A ROMPC formulation that can effectively capture residual error dynamics with low data requirements could also be leveraged to address the problem of simulation-to-reality transfer---a challenge that is necessary to address for most modern robotic systems. Similarly, adaptive control frameworks could be developed that leverage streaming-based methods for updating ROMs during operation of the robot, allowing an initial model to be trained in simulation and updated after transferring to the physical system.  

\section*{Acknowledgments}
I.A., M.P., J.C., and M.T.T.  were supported by the U.S. Office of Naval Research (ONR), grant no. N00014-22-1-2595. B.K. and H.S. were supported by ONR under award number N00014-22-1-2624. Any opinions, findings, and conclusions or recommendations expressed in this material are those of the authors and do not necessarily reflect the views of the ONR.


\bibliographystyle{ieeetr}
\bibliography{references_open} 

\begin{IEEEbiography}[{\includegraphics[width=1in,height=1.25in,clip,keepaspectratio]{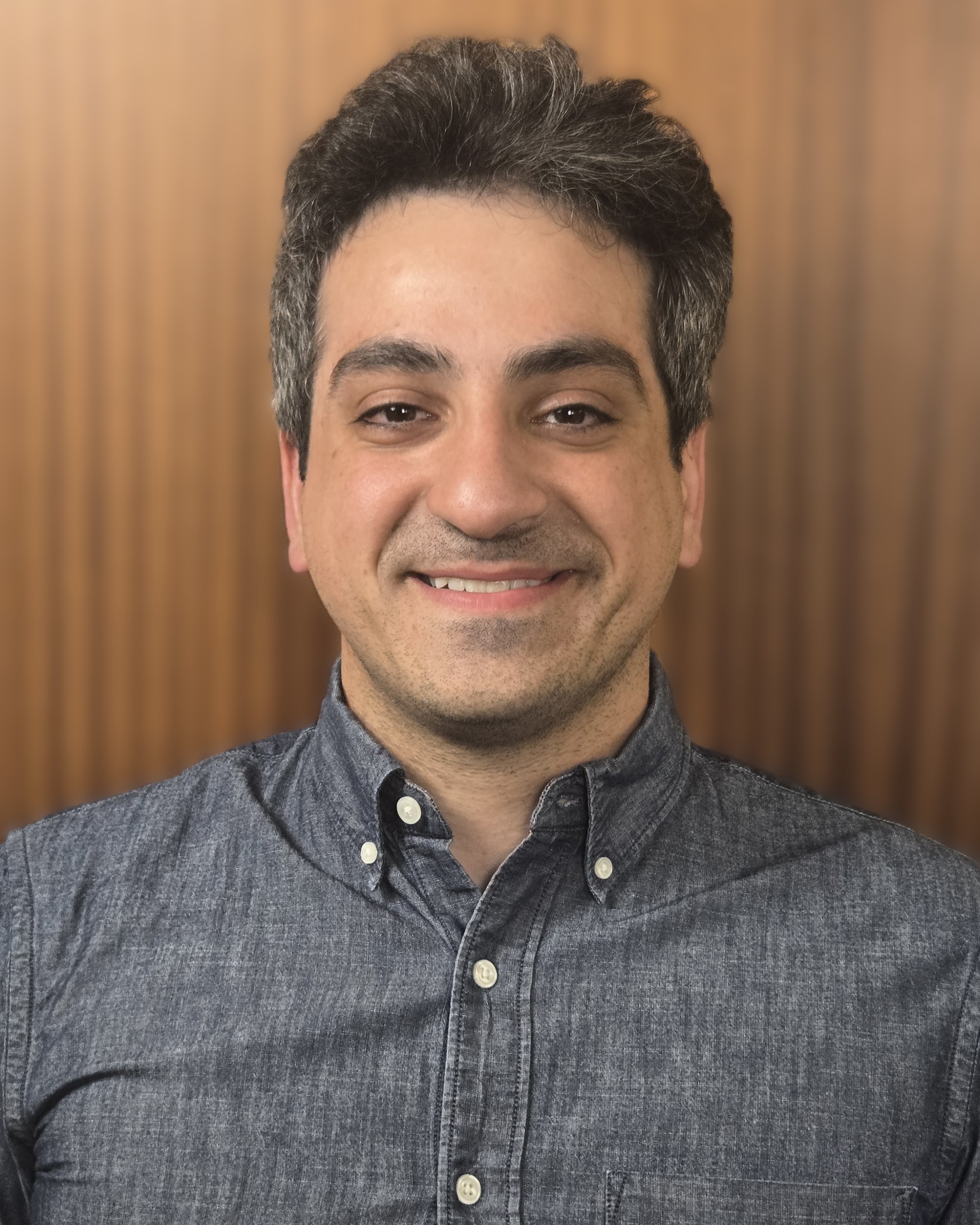}}]{Iman Adibnazari} (Member, IEEE)
is a Ph.D. candidate in the Department of Mechanical and Aerospace Engineering at the University of California San Diego, CA, USA. Previously he received the B.S. degree in electrical engineering from the University of Utah, Salt Lake City, UT, USA, in 2018 and M.S. in electrical and computer engineering from the University of California San Diego, CA, USA, in 2020. 

His research interests include design and fabrication of robotic systems, data-driven methods for modeling and control, and robot ethics. 

He was recognized as a Science Policy Fellow from 2022-2025 from the School of Global Policy and Strategy at the University of California San Diego. 
\end{IEEEbiography}

\begin{IEEEbiography}[{\includegraphics[width=1in,height=1.25in,clip,keepaspectratio]{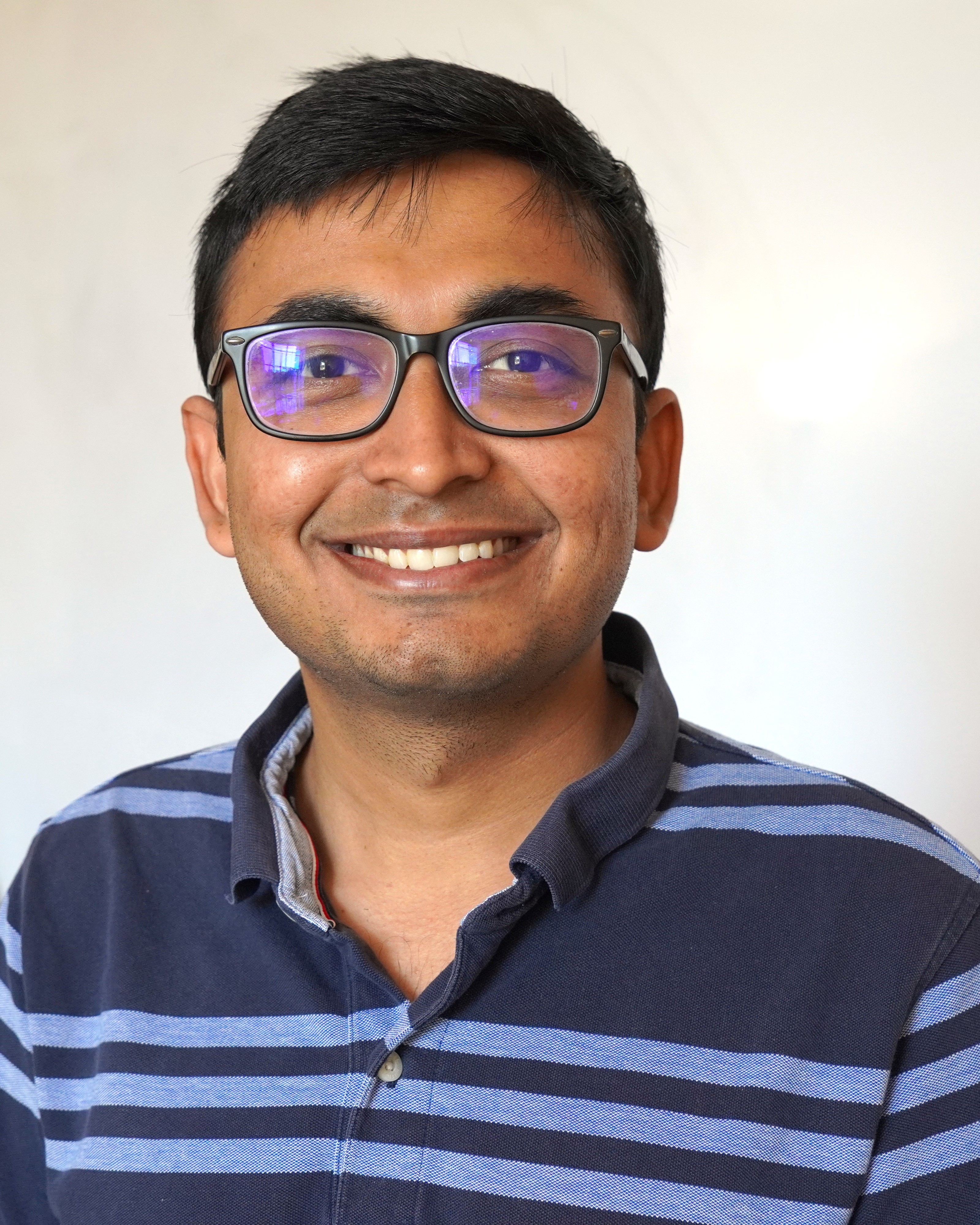}}]{Harsh Sharma} (Member, IEEE)
is an Assistant Professor in the Department of Mechanical Engineering at the University of Wisconsin–Madison, WI, USA. Prior to this, he was a Postdoctoral Research Scholar at the University of California, San Diego, CA, USA. He received the Ph.D. degree in Aerospace Engineering and the M.S. degree in Mathematics from Virginia Tech, Blacksburg, VA, USA, and the B.Tech. and M.Tech. degrees in Mechanical Engineering from the Indian Institute of Technology Bombay, Mumbai, India. 

His research interests include reduced-order modeling, scientific machine learning, and structure-preserving methods for the design, analysis, and control of complex dynamical systems, with applications in computational physics, structural dynamics, and soft robotics.
\end{IEEEbiography}

\begin{IEEEbiography}[{\includegraphics[width=1in,height=1.25in,clip,keepaspectratio]{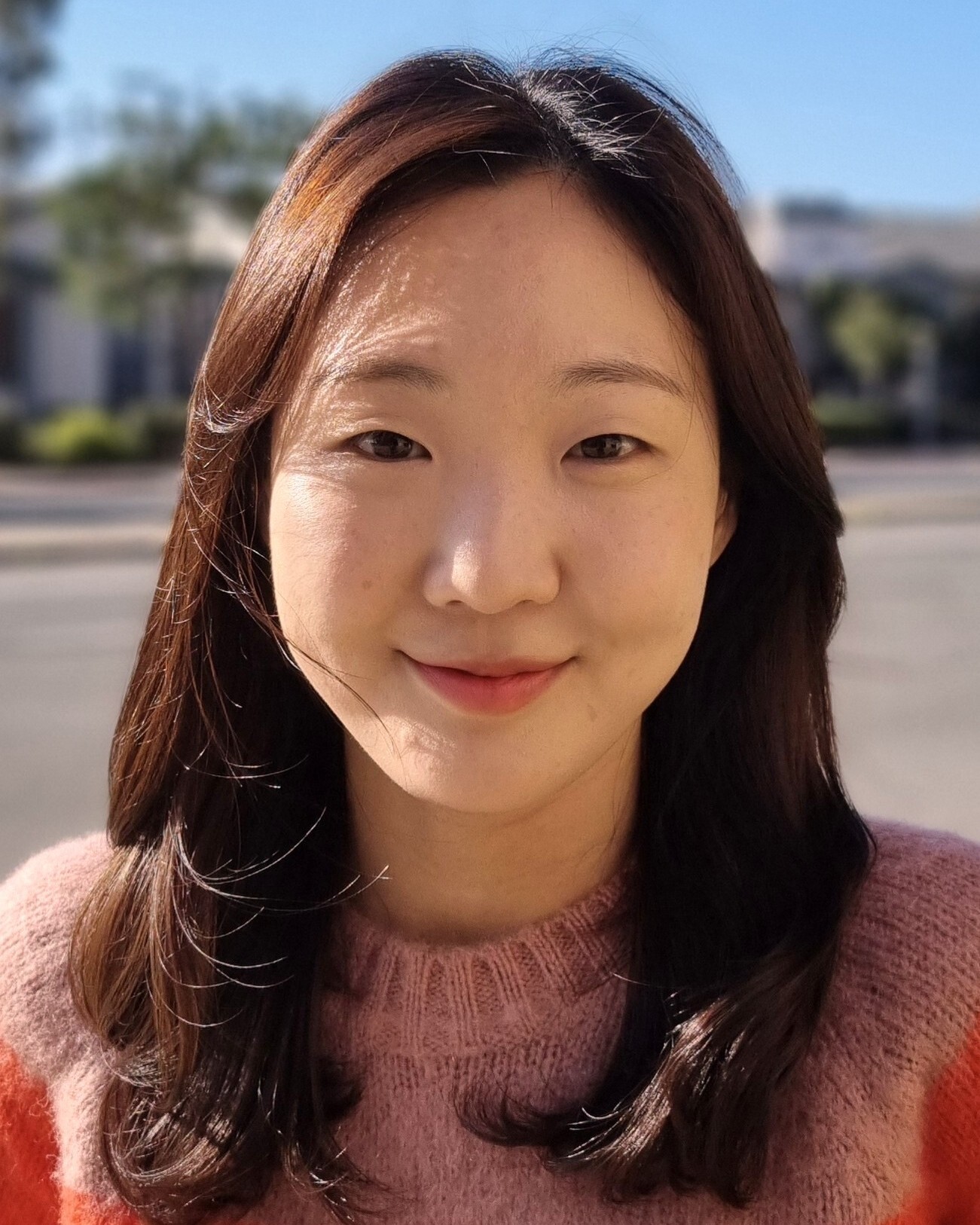}}]{Myungsun Park}
(Member, IEEE) is a Post-doctoral researcher at University of California San Diego. She received the Ph.D. degree in 2024 from Seoul National University, Korea. 

Her research interests include hand motion tracking, control of robot hands, development of soft sensors for proprioceptive, tactile and environmental sensing, and underwater robotic systems.
\end{IEEEbiography}

\begin{IEEEbiography}[{\includegraphics[width=1in,height=1.25in,clip,keepaspectratio]{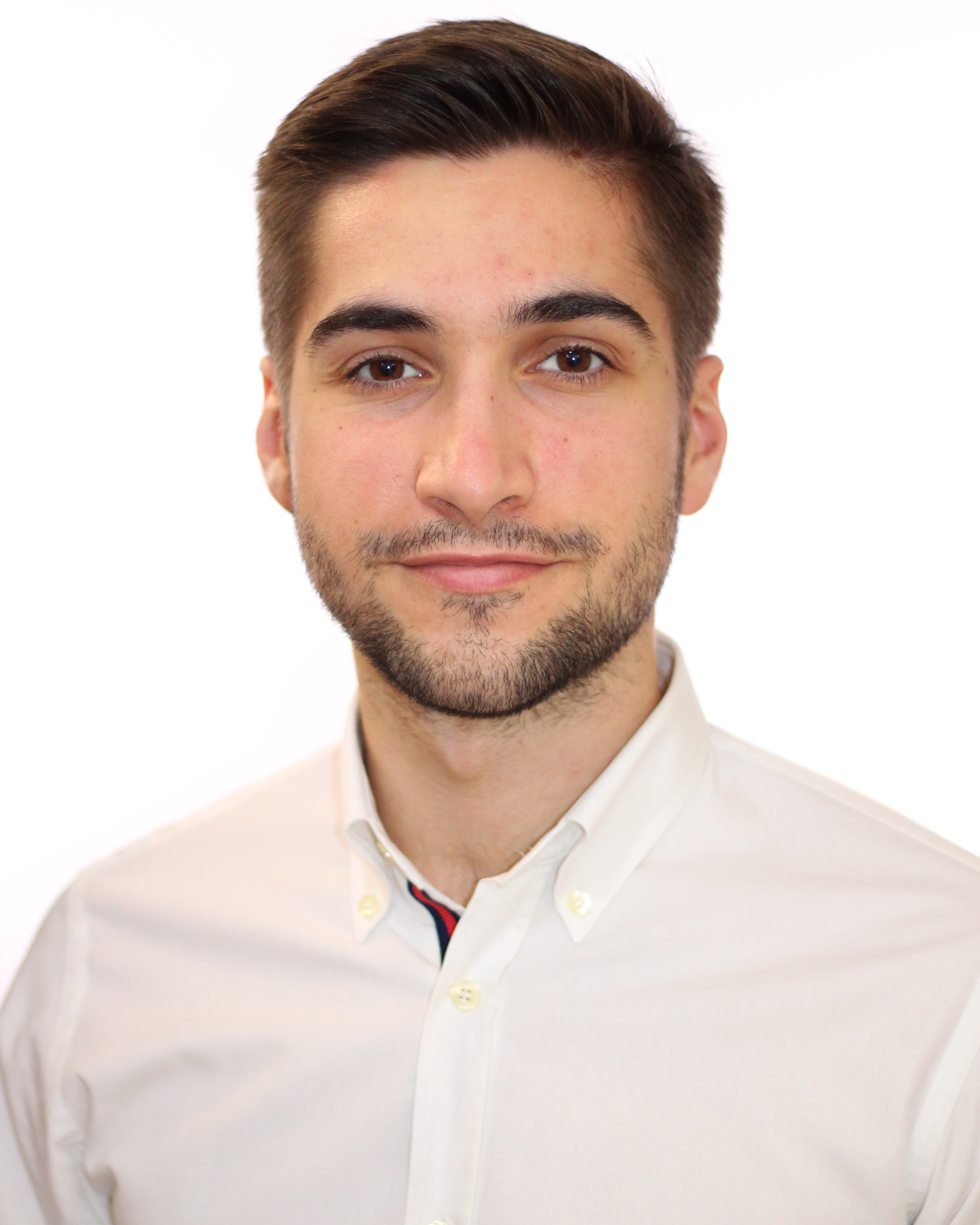}}]{Jacobo Cervera-Torralba}
(Graduate Student Member, IEEE) received the B.Sc. degree in aerospace engineering from the Polytechnic University of Valencia, Valencia, Spain, in 2021. He is currently working toward the Ph.D. degree in design and fabrication of bio-inspired soft robots for underwater exploration with the Department of Mechanical and Aerospace Engineering, University of California, San Diego, San Diego, CA, USA. 

His research focuses on the design and fabrication of bioinspired soft robots for underwater exploration. 

Mr. Cervera-Torralba was the recipient of the Charles Lee Powell Fellowship 2021.
\end{IEEEbiography}

\begin{IEEEbiography}[{\includegraphics[width=1in,height=1.25in,clip,keepaspectratio]{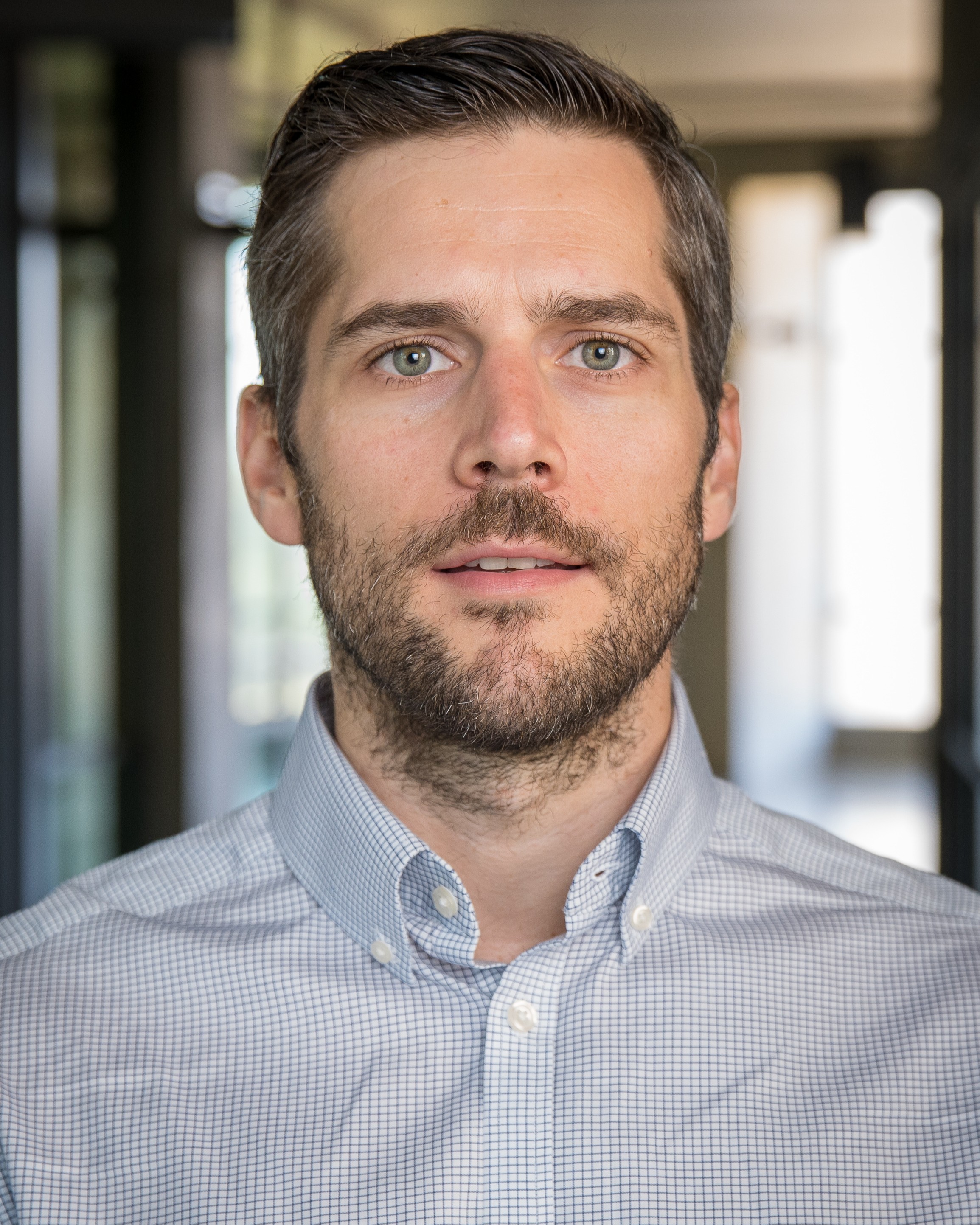}}]{Boris Kramer}
(Member, IEEE) received the M.S. and Ph.D. degrees in mathematics from Virginia Tech, Blacksburg, Virginia, in 2011 and 2015, respectively. 

He is an Associate Professor in Mechanical and Aerospace Engineering with the University of California San Diego, San Diego, CA, USA. From 2015 to 2019, he has been a Postdoctoral Scholar with the Massachusetts Institute of Technology, Cambridge, MA, USA. His main research interests include model reduction, high-dimensional control, data driven modeling, (multifidelity) uncertainty quantification, and design under uncertainty.

 Dr. Kramer is a Member of SIAM and a Senior Member of AIAA. He was the recipient of the National Science Foundation Early CAREER Award in Dynamics, Control, and System Diagnostics in 2022 and the Department of Defense Newton Award in 2020. He is currently an Associate Editor for the SIAM/ASA Journal on Uncertainty Quantification.\end{IEEEbiography}

 \begin{IEEEbiography}[{\includegraphics[width=1in,height=1.25in,clip,keepaspectratio]{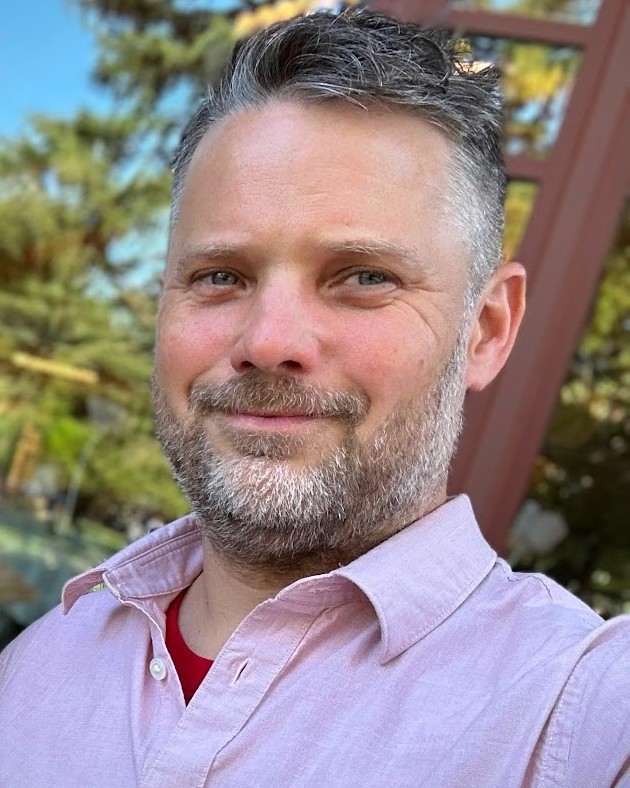}}]{Michael T. Tolley}
(Senior Member, IEEE) received the Ph.D. and M.S. degrees in mechanical engineering with a minor in computer science from Cornell University, Ithaca, NY, USA, in 2009 and 2011, respectively.

He is currently a Professor in Mechanical and Aerospace Engineering, and the Director of the Bioinspired Robotics and Design Lab, Jacobs School of Engineering, University of California San Diego (UCSD), San Diego, CA, USA. Before joining the Mechanical Engineering faculty with UCSD in the fall of 2014, he was a Postdoctoral Fellow with the Wyss Institute for Biologically Inspired Engineering, Harvard University, Cambridge, MA, USA. His research seeks inspiration from nature to design robotic systems with the versatility, resilience, and efficiency of biological organisms. 

Dr. Tolley’s work has appeared in leading academic journals including Science and Nature, and has been recognized by awards including a U.S. Office of Naval Research Young Investigator Program award and a 3 M Nontenured Faculty Award.
\end{IEEEbiography}

\vfill

\end{document}